\pdfoutput=1
\documentclass[11pt]{article}
\usepackage[final]{acl}
\usepackage{times}
\usepackage{latexsym}
\usepackage{enumitem}
\usepackage[breakable]{tcolorbox}
\usepackage[T1]{fontenc}
\usepackage[utf8]{inputenc}
\usepackage{xcolor}
\usepackage{amsmath}
\usepackage{multirow}
\usepackage{microtype}
\usepackage{tikz}
\usepackage{inconsolata}
\usepackage{makecell}
\usepackage{booktabs}
\usepackage{graphicx}
\usepackage{cleveref}
\definecolor{promptpurple}{RGB}{147,51,147}
\definecolor{instructionframe}{RGB}{51, 122, 183}
\definecolor{instructionback}{RGB}{240, 247, 255}
\definecolor{observationframe}{RGB}{70, 136, 71}
\definecolor{observationback}{RGB}{223, 240, 216}
\definecolor{demoframe}{RGB}{128, 0, 128}
\definecolor{demoback}{RGB}{245, 235, 255}
\crefname{figure}{Figure}{Figures}
\crefname{table}{Table}{Tables}
\crefname{appendix}{Appendix}{Appendices}
\crefname{section}{Section}{Sections}
\crefname{equation}{Eq.}{Eqs.}
\crefname{enumi}{}{} 
\usepackage{pifont}
\newcommand{\ccirc}[1]{\textcolor{#1}{\ding{108}}}
\definecolor{left_right_analysis_nude}{HTML}{cdb39c}
\definecolor{left_right_analysis_orange}{HTML}{e86100}
\definecolor{left_right_analysis_green}{HTML}{00bb77}

\renewcommand\paragraph[1]{\noindent{\bf {#1}. }}
\newcommand*\samethanks[1][\value{footnote}]{\footnotemark[#1]}

\newif\ifarxiv
\arxivfalse

\title{DARS: Dynamic Action Re-Sampling to Enhance Coding Agent Performance by Adaptive Tree Traversal}

\author{%
  Vaibhav Aggarwal\thanks{{ } Co-first author. 
  \quad \quad $^\dag$Co-supervision.}
  \\
   Sprinklr \\
  \texttt{ \href{mailto:vaibhav\_a@ee.iitr.ac.in}{vaibhav\_a@ee.iitr.ac.in}} 
  \And
  Ojasv Kamal\samethanks{} 
  \\
  Mirelo AI \\
  \texttt{ \href{mailto:kamalojasv2000@gmail.com}{kamalojasv2000@gmail.com}} \\
  \And
  Abhinav Japesh
  \\
  Sprinklr\\
  \texttt{ \href{mailto:akjapesh@gmail.com}{akjapesh@gmail.com}} \\
  \AND
  Zhijing Jin$^\dag$ 
  \\
  MPI-IS \& University of Toronto \\
  \texttt{\href{mailto:zjin@cs.toronto.edu}{zjin@cs.toronto.edu}} \\
  \And
  Bernhard Schölkopf$^\dag$
  \\
  MPI for Intelligent Systems \\
  \texttt{\href{mailto:bs@tue.mpg.de}{bs@tue.mpg.de}} \\
}

\begin{document}
\maketitle
\begin{abstract}
Large Language Models (LLMs) have revolutionized various domains, including natural language processing, data analysis, and software development, by enabling automation. In software engineering, LLM-powered coding agents have garnered significant attention due to their potential to automate complex development tasks, assist in debugging, and enhance productivity. However, existing approaches often struggle with sub-optimal decision-making, requiring either extensive manual intervention or inefficient compute scaling strategies. To improve coding agent performance, we present Dynamic Action Re-Sampling (DARS), a novel inference time compute scaling approach for coding agents, that is faster and more effective at recovering from sub-optimal decisions compared to baselines. While traditional agents either follow linear trajectories or rely on random sampling for scaling compute, our approach DARS works by branching out a trajectory at certain key decision points by taking an alternative action given the history of the trajectory and execution feedback of the previous attempt from that point. We evaluate our approach on SWE-Bench Lite benchmark, demonstrating that this scaling strategy achieves a pass@k score of 55\% with Claude 3.5 Sonnet V2. Our framework achieves a pass@1 rate of 47\%, outperforming state-of-the-art (SOTA) open-source frameworks.\footnote{Our codes 
are at \url{https://github.com/darsagent/DARS-Agent}, datasets and models at \url{https://huggingface.co/AGENTDARS}, and a demo of our trajectory analysis tool at \url{https://darsagent.github.io/DARS-Agent/}}

\end{abstract}

\section{Introduction}
\label{introduction}

Software engineering (SWE)  has become increasingly critical in modern technology development, with developers spending countless hours writing, reviewing, and maintaining code, creating an urgent need for automation to improve productivity \cite{swesurvey}. Large language models (LLMs) have emerged as promising tools for automating various software engineering tasks, with breakthrough works like SWE-bench \cite{swebench} establishing evaluation frameworks and datasets, leading to widespread adoption of tools such as \citet{sweagent,opendevin}.

There are three primary approaches to developing SWE agents based on LLMs. The first follows a sequential ReAct \cite{react} loop, where agents such as SWE-Agent \cite{sweagent} and OpenDevin \cite{opendevin} interact with development tools and incorporate execution feedback to refine their predictions. The second approach generates multiple candidate solutions using temperature-based sampling and then selects the best one through ranking \cite{masai} or majority voting \cite{agentless}. The third approach, exemplified by SWE-Search \cite{swe_search}, leverages Monte Carlo Tree Search (MCTS) \cite{uct} to systematically explore the solution space.

However, each method has limitations: (1) Sequential agents struggle to recover from suboptimal decisions due to context length constraints \cite{babilong,longcontext}. (2) Multi-solution approaches lack efficient mechanisms for knowledge sharing between independently generated solutions. (3) Tree search methods, such as SWE-Search \cite{swe_search}, rely on scalar value functions and suffer from slow execution speeds, making them less effective for long-horizon planning.

To address these challenges, we propose Dynamic Action Re-Sampling (DARS), which enhances coding agents by dynamically re-sampling actions based on prior execution results. Instead of generating multiple independent trajectories, DARS selectively branches at key decision points, using a depth-first strategy to fully explore a trajectory before branching. This offers two advantages: a) \textbf{Long Horizon Feedback}: Our experiments show improved pass@1 rates \ref{tab:lookahead} by providing complete trajectory feedback before branching.  b) \textbf{Efficiency}: Depth-first search reduces memory overhead by reusing the current environment state without simulating future states. 
Finally, we introduce a trajectory selection pipeline leveraging proprietary and preference-optimized models to identify the most promising solution \cite{prometheus2}.

Across the experiments, our DARS method achieves up to 47\% pass@1 rate, which is open-source SOTA performance on the SWE-Bench Lite benchmark \cite{swebench}.

In conclusion, the main contributions of our approach are as follows:
\begin{enumerate}[nolistsep]
    \item We introduce DARS, an inference-time compute scaling method for coding agents that rapidly recovers from suboptimal decisions, achieving an open source SOTA pass@1 rate of 47\% on the SWE-Bench Lite benchmark.
    \item We propose a patch preference data generation and supervised fine-tuning pipeline to select the most promising solution among multiple attempts.
    \item We release our complete codebase, a 500M-token execution feedback critique dataset, model checkpoints (7B, 14B, and 32B), and a trajectory analysis tool to support future research.
\end{enumerate}

\section{Related Work}
\label{related_work}
\vspace{-0.25cm}
\paragraph{LLM Agents for Software Engineering}
\label{related_work__coding_agents}
Large Language Model (LLM) agents have been increasingly employed to automate software engineering tasks such as bug fixing and code generation. These agents integrate tools for code editing, search, navigation, and execution \citep{sweagent}. Enhancements in this domain include diff-based editing \citep{aider}, execution with Jupyter and web search capabilities \citep{opendevin}, and optimized repository search \citep{aider,autocoderover,RepoGraph,aorwall_moatless_2024}. Some approaches further modularize functionalities to improve efficiency \citep{agentless,masai}.

Recent studies have explored generating multiple solutions to enhance accuracy. For instance, \citet{llmmonkeys} demonstrated that sampling 250 solutions can increase accuracy by 250\%. However, methods like those proposed by \citet{agentless} and \citet{masai} rely on inefficient random sampling. To address this, \citet{swe_search} introduced an approach that improves efficiency through Monte Carlo Tree Search (MCTS) \citep{uct,mcts_1}, balancing computational resources with scalar rewards and textual feedback. Despite these advancements, their reliance on retrospective feedback limits early guidance, and frequent environment resets can slow execution.

DARS improves efficiency by branching only at critical decisions and providing long-horizon feedback, reducing resets and accelerating execution.

We discuss about Inference Time Compute Scaling and LLM as Code Reviewers in \cref{sec:appendix_more_related_works}

\begin{figure*}[t]
    \centering
    \includegraphics[width=\textwidth]{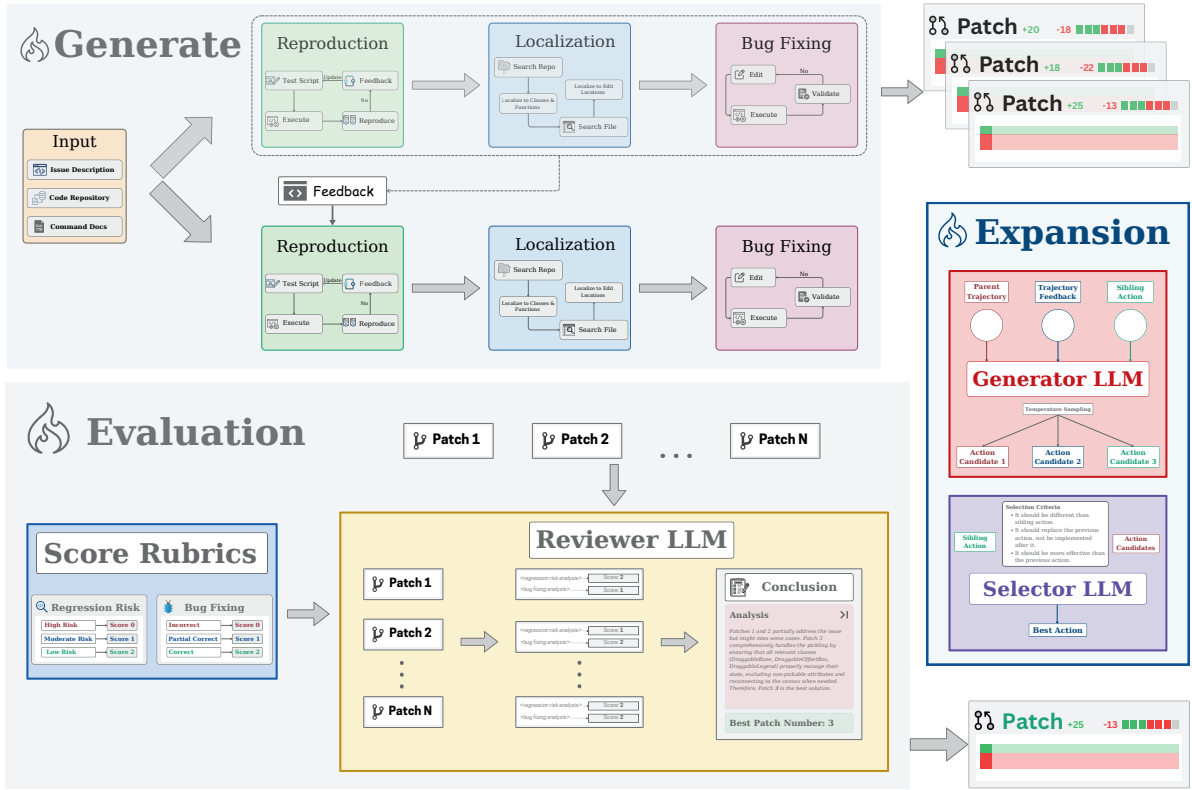}
    \caption{
        Overview of our DARS scaling method. 
        DARS processes issue-related information and generates multiple patches using the \textbf{Expansion} mechanism. 
        These patches are then evaluated by our \textbf{Reviewer LLM}, which assigns scores based on predefined \textbf{Score Rubrics}, ultimately selecting the best patch for output.
    }
    \label{fig:main_fig}
\end{figure*}
\vspace{-0.25cm}

\section{Our DARS Method}
\vspace{-0.3cm}
\label{approach}
The main motivation behind DARS is to enhance the agent's ability to recover and learn from sub-optimal decisions by taking alternative actions while minimizing redundancy. However, errors also scale with scaling trajectories, therefore, we optimize our backbone SWE-Agent first by improving its editing capabilities and adding various actions to it. We then identify the most promising action types by optimizing the trade-off between increase in resolve rate and increase in cost due to branching the trajectory at that point. We define this process of branching the tree as expansion. We finally select the most promising trajectory from all the attempts. We go over the details of each of these steps in the following sections.

\subsection{Improving the Base SWE Agent}
\label{approach__base_agent}
\subsubsection{Editing Capabilities}
We build on the SWE-Agent \cite{sweagent} which uses a ReAct loop to iteratively generate a thought and an action and receive feedback from the sandbox environment. By default, the agent uses a whole style of editing where it needs to generate the start and end line numbers of the edit followed by the content of edit. This type of edit often results in numerous syntax and semantic errors (see Section \ref{sec:imp_editing}), as the agent fails to account for both the targeted and adjacent code, leading to issues such as indentation errors. We enhance the editing process using Aider \cite{aider}, a diff-based tool. In this approach, the agent generates both the content to be replaced and its replacement. Additionally, to facilitate content addition without replacement, Aider introduces two new actions: append and insert. 

\noindent \textbf{Append} adds content at the end of a file, while \textbf{Insert} requires a line number and the content to be inserted at that specific location. This approach compels the model to better consider the existing code. Moreover, we enhance the editing process by having the agent output both the \texttt{to\_replace} and \texttt{replace\_with} contents, each followed by a \texttt{\$} character to properly escape special characters (e.g., newlines, quotes) as described in Section~\ref{tool instructions}.
\vspace{-0.25cm}
\subsubsection{New Actions}
In addition to editing, we introduce several new actions to enhance each of the three stages of our bug-fixing process. We add the following actions to the agent:

\paragraph{Execute Server} The sandbox environment limits the execution of iterative or long-running scripts. To address this, we add the \emph{execute server} action with persistent memory. Instead of retrieving code output directly, the agent uses \texttt{get\_logs} to access execution logs. This action is especially effective during the reproduction stage for efficiently replicating bugs (see Section \ref{tool instructions}).

\paragraph{Execute IPython} This action enables the agent to run Python code within an IPython environment, streamlining bug reproduction by eliminating the need to create, write, and execute a separate file.
\vspace{-0.1cm}
\paragraph{Search Repo} Search repo command uses a cached RepoGraph \cite{RepoGraph}—a hierarchical structure where nodes represent code definitions and edges represent dependencies between them. By utilizing sub-graph retrieval algorithms, RepoGraph extracts ego-graphs \cite{ego-graph} centered around specific keywords. This action allows the agent to search for a specific keyword in the repo and get all the files with corresponding line numbers where the keyword is present, to aid in precise bug localization.
\vspace{-0.1cm}
\paragraph{Undo Edit} Often times, the agent makes a mistake in the edit and needs to undo it. However, due to inherent limitations of editing abilities of the agent, the agent sometimes outputs syntactically incorrect code which degrades its reasoning \cite{babilong} capabilities and takes up its computational budget. This action allows the agent to directly undo the last edit efficiently \cite{anthropic_quickstarts}

\subsection{DARS Scaling}
\label{approach__dars}
DARS begins by completing a trajectory in a depth-first manner while storing key decision nodes in a priority queue, sorted in ascending order by node depth. Once the current trajectory reaches a terminal state—either by a submit command (see \ref{backbone_prompts}) or upon reaching a predefined maximum depth—these nodes are expanded. During expansion, we sample $k$ alternative actions and select the best one. We define key decision points as those actions that significantly enhance the resolve rate at minimal cost. As shown in \cref{tab:action_model_comparison}, expanding the trajectory at edit actions is particularly effective. This approach allows the agent to learn from previous mistakes and recover from suboptimal decisions, which is crucial for long-horizon tasks like programming. Finally, if no branch submits code before reaching the maximum depth, the code is auto-submitted. Issues that fail to execute the expected trajectory due to runtime errors or other anomalies within the SWE-Agent environment are re-run.

\subsubsection{Branching Strategy}
\label{approach__dars__branching_strategy}
The main improvement in DARS lies in the avoidance of branching out trajectory at all actions, which costs exponential compute and its redundancy leads to a low accuracy for the trajectory selection pipeline. 
We use a causal analysis in \cref{tab:action_model_comparison} to identify four key actions with the largest causal impact on the model performance: edit, append, create, and submit and further perform qualitative analysis in \cref{sec_expansion_qualitative_anal} to understand the reasons behind the results. 

\paragraph{Create} 
Reproduction scripts are essential for debugging. Insufficient details can hinder their effectiveness and lead to incorrect fixes. By first localizing and analyzing the relevant code, the model improves bug resolution.

A key issue is that models often fail to refine reproduction scripts during bug fixing. While some cases improve, others show overconfidence, with flawed scripts being repeated. Prioritizing localization is crucial for accurate reproduction.

The \texttt{create} action differs from \texttt{append} in reproduction scripts. Though both evaluate and fix bugs, \texttt{append} actions generally produce better scripts. Models are often biased by previous actions, making only minor script changes instead of exploring new paths. Early sampling of solutions during \texttt{create} allows better exploration (see ref \ref{fig:create_better}).

Should issue localization always precede reproduction? Not always. Early localization can bypass reproduction, leading to weaker solutions or misinterpretations of the bug, as shown in \cref{fig:create_worst}. Reproducing the bug first enables a clearer understanding and more accurate fixes.

\paragraph{Append} The append action improves reproduction scripts by refining previous attempts, ensuring tests sufficiently verify code edits (see \ref{fig:append_better_rep})

Runtime errors arise when the model lacks codebase or environment knowledge, hindering issue reproduction and exhausting its reasoning context. Expansion in append actions mitigates this by accelerating the reproduction phase, reducing turns needed for localization. This allows more iterations for editing and testing, improving bug resolution. The benefit occurs in two ways: direct bug identification during expansion or improved reproduction scripts enabling better localization. (see \ref{fig:append_better_loc})

\paragraph{Edit} The agent sometimes generates semantically incorrect code, leading to an edit-Python loop. As context length grows, its reasoning weakens, trapping it in an unproductive cycle without a clear exit (see \ref{fig:edit_python_loop})

The agent frequently produces code with basic syntax errors, such as mismatched parentheses or incorrect indentation, leading to a cycle of repetitive fixes. Due to reasoning flaws, it often gets stuck applying the same ineffective edits—such as repeatedly adding a closing bracket—even when the fix has already failed (see 
 \ref{fig:edit_loop}).

\paragraph{Submit} While the model can fix bugs, it sometimes introduces regressions. To prevent this, it should verify changes by running tests and refining edits based on results. Expansion in this action prompts the model to reassess its fixes and correct issues before submission. For example, in the following case, the model fixes a bug but introduces a regression. By reevaluating its changes, it catches and resolves the issue (see \ref{fig:submit_better}).

We further cut down the redundancy by considering second-degree expansions. The above actions namely create, append, edit, and submit usually occur in the same order. The higher the action in the tree, the higher the impact of expanding the tree at that action. Therefore, if a branch is expanded at create, we only expand the tree at append, edit, and submit the next time. Similarly, a branch expanded at edit is only expanded at submit the next time. We follow this rule with an exception in the case of append, since empirically it has found that this has led to a high resolve rate for the extra cost incurred. Finally, for each branch, we put a cap on the number of expansions of each type to prevent the tree from growing exponentially.
\vspace{-0.3cm}
\subsubsection{Expansion Strategy}
\label{approach__dars__expansion_strategy}
We use a depth-first strategy to explore the current trajectory before branching out, which has two main advantages: \textbf{Speed} and \textbf{Long-Horizon Feedback}. After reaching a terminal condition, we continue from the node with the lowest depth in the priority queue. In \cref{fig:traversal-strategy}, we find that lowest depth-first is the most effective strategy, as the flexibility to explore decreases with node depth.

\vspace{-0.25cm}
\subsection{Best Trajectory Selection}
After the agent has generated multiple trajectories, we select the most promising trajectory from all the attempts in two stages namely trajectory pruning and trajectory selection. We begin by cleaning the patches submitted by each trajectory by removing any bug reproduction files from it. We then prune any redundant trajectories which lead to the same cleaned patch. In the second stage, we use off-the-shelf open and close source models as well as our custom supervised-fine-tuned models to choose the best trajectory based on custom rubrics namely reproduction, fix, and potential to introduce new bugs motivated by \cite{prometheus2}.

\label{approach__trajectory_selection__training_details}
\paragraph{Patch Preparation} We begin with cleaning the patches by removing everything except the bug fix part. This includes removing the bug reproduction script, readme / documentation changes, pycache files etc. We then generate critiques for each patch based on the three rubrics namely reproduction, fix, and potential to introduce new bugs. To ground the predictions of the model, we use the execution output obtained from running the tests after applying the patch.

\paragraph{Patch Sampling} For a given issue, based on the distribution of number of patches generated by DARS for that issue, we sample all combinations of patches from 2 to 6 patches. We further do a fine-grained sampling of negative patches by dividing all the negative patches in buckets based on the combinations of tests that fail after applying the patch, to get a balanced dataset. For positive patches, we sample from the set of all the positive patches if there are any. For cases where there are no positive patches, we just use the ground truth patch.

\section{Experimental Setup}
\label{experimental_setup}

\begin{table}[ht]
    \setlength{\tabcolsep}{1pt} 
    \begin{center}
    \begin{small}     \centering
    \begin{tabular}{llccccrr}
    \hline
    Framework &  Base Model & Pass@1 \\
    \hline
    SWE-Agent & GPT-4o & 18.3  \\
    SWE-Agent & Claude 3.5 Sonnet V2 & 23.0  \\
    Moatless Tools & GPT-4o & 24.7  \\
    Aider & GPT-4o \& Claude 3 Opus & 26.3  \\
    Moatless Tools & Claude 3.5 Sonnet V2 & 38.3  \\
    MASAI & GPT-4o & 27.3  \\
    Agentless-1.5 & GPT-4o & 27.3  \\
    Moatless Tools & Claude 3.5 V2 & 38.3  \\
    Agentless-1.5 & Claude-3.5 V2 & 40.7  \\
    OpenHands & CodeAct v2.1 & 41.7  \\
    SWE-Search & GPT-4o & 31.0  \\
    Kodu-v1 & Claude-3.5 Sonnet V2 & 44.7  \\
    \textbf{DARS (Ours) } & Claude 3.5 Sonnet + Deepseek R1 & \textbf{47.0}  \\
    \hline
    \end{tabular}
    \end{small}
    \end{center}
    \caption{Comparative analysis of various software engineering agents' performance on SWE-Bench Lite dataset. We present results only for the language models that were used by the respective authors, as evaluating every possible combination of models and frameworks is highly resource-intensive.}
    \label{tab:framework_comparison}
    \vskip -0.1in
\end{table}
\vspace{-0.2cm}
\subsection{DARS Scaling}

\paragraph{Dataset}
We use the SWE-Bench Lite benchmark, a widely used subset of the SWE-Bench dataset \cite{swebench}. It comprises 300 GitHub issues from 12 real-world software projects, each containing an issue report and the corresponding codebase.

\begin{table}[ht]
    \setlength{\tabcolsep}{1pt} 
    \begin{center}
    \begin{small}     \centering
    \begin{tabular}{lcccccccc}
    \hline
    & \multicolumn{2}{c}{7B} & \multicolumn{2}{c}{14B} & \multicolumn{2}{c}{32B} & \\
    \cmidrule(r){2-3} \cmidrule(r){4-5} \cmidrule(r){6-7}
    & Vanilla & FT & Vanilla & FT & Vanilla & FT & R1 \\
    \hline
    GPT-4o & 33.0 & 33.3 & 35.7 & 37.0 & 36.0 & 36.7 & 37.0 \\
    Gemini-1.5-pro & 26.3 & 26.3 & 27.0 & 27.7 & 29.7 & 29.0 & 33.0 \\
    Gemini-2.0-flash & 26.7 & 27.0 & 26.3 & 27.7 & 28.0 & 28.7 & 28.3 \\
    Claude 3.5 Sonnet & 35.7 & 38.7 & 39.7 & 41.7 & 41.3 & 42.0 & \textbf{47.0} \\
    \hline
    \end{tabular}
    \end{small}
    \end{center}
    \caption{Performance Comparison across 7B, 14B, and 32B parameter DeepSeek R1 Distill Qwen reviewer models}
    \label{tab:reviewer_performance_comparison}
    \vskip -0.1in
\end{table}

\paragraph{Evaluation Metrics}
We evaluate model performance using multiple metrics. \textbf{Resolve Rate} (Pass@1) measures the fraction of instances fixed on the first attempt, while \textbf{Pass@k} represents the expected success rate within $k$ attempts. To assess efficiency, we track the \textbf{Average Cost per Instance} (in dollars) and the \textbf{Cost Scaling Factor}, which compares scaled resource costs to the base agent. Lastly, we record the \textbf{Number of Attempts} required for a successful fix.

\begin{table}[ht]
    \vskip 0.15in
    \centering
    \begin{small}
    \begin{tabular}{lccc}
    \hline
    Method & Overall & Pass@1 & Precision \\
    \hline
    Complete & 10 & 9 & 0.51 \\
    5 look aheads & 5 & 2 & 0.54 \\
    10 look aheads & 8 & 5 & 0.68 \\
    Path Summary & 9 & 5 & 0.54 \\
    Only Sibling Action & 8 & 5 & 0.72 \\
    \hline
    \end{tabular}
    \end{small}
    \caption{Variation of performance with horizon of context during expansion.}
    \label{tab:lookahead}
    \vskip -0.1in
\end{table}

\paragraph{Baselines}
We test our approach against various SWE agents including SWE-Agent \cite{sweagent}, Moatless Tools \cite{aorwall_moatless_2024}, and OpenHands \cite{opendevin}, MASAI \cite{masai}, Large Language Monkeys \cite{llmmonkeys}, Agentless \cite{agentless}, and SWE-Search \cite{swe_search}. In terms of LLMs, we test our approach with various models including GPT-4o \cite{gpt4o}, Claude 3.5 Sonnet V2 \cite{claude35sonnet}, Gemini 2.0 Flash, and Gemini 1.5 Pro \cite{team2023gemini}.

\begin{table*}[!htbp]
    \vskip 0.15in
    \begin{center}
    \begin{small}     \centering
    \begin{tabular}{@{}l@{\hspace{6pt}}l@{\hspace{4pt}}c@{\hspace{4pt}}c@{\hspace{4pt}}c@{\hspace{4pt}}c@{\hspace{4pt}}c@{\hspace{4pt}}c@{}}
    \hline
    Framework & Model & \makecell[c]{Cost Scaling Factor} & \makecell[c]{\# Attempts} & \makecell[c]{Single Rollout } & \makecell[c]{Coverage } & \makecell[c]{$\Delta$} & \makecell[c]{Precision } \\
    \hline
    Agentless & GPT-4o & -- & 40 & -- & 42 & -- & -- \\
    MASAI & GPT-4o & -- & 5 & 23 & 35 & 34.28 & -- \\
    {Large Language Monkeys} & 
    {DeepSeek-Coder} & 250 & 250 & 15.9 & 56 & 71 & 14 \\
    SWE-Search & GPT-4o & 14.00 & 5.00 & 25.70 & 34.00 & 24.41 & 20.00 \\
    \textbf{DARS (Ours) }& GPT-4o & 7.60 & 5 & 21.67 & 43.34 & 50.00 & 75.00 \\
    \hline
    \end{tabular}
    \end{small}
    \end{center}
    \caption{Compute Scaling Efficiency comparison across various frameworks and metrics. Here Single Rollout represents the performance of the agent when a single trajectory is generated.}
    \label{tab:compute-scaling-efficiency}
    \vskip -0.1in
\end{table*}

\paragraph{Hyperparameters}
The DARS algorithm relies on several key hyperparameters. \emph{Num Expansions} is set to 2, defining the number of expansions per decision point, while \emph{Expansion Temperature} (0.8) controls the sampling temperature for alternative actions. The algorithm runs for 300 iterations (\emph{Num Iterations}), with a maximum branch depth of 50 (\emph{Max Branch Depth}). Action limits are defined by \emph{Expansion Limit Edit, Append, Submit,} and \emph{Create}, each set to 1 which caps the number of times those actions can be expanded within a branch. \emph{Num Expansion Sampling} (3) specifies the number of sampled actions per expansion, and \emph{Num Lookahead} (50) determines how many steps from previous trajectory are considered during tree expansion.

\begin{table*}[t]
    \vskip 0.15in
    \begin{center}
    \begin{small}     \centering
    \begin{tabular}{lccccccc}
    \toprule
    & \multicolumn{2}{c}{SWE-Agent} & \multicolumn{2}{c}{Improved SWE-Agent} & \multicolumn{2}{c}{DARS} \\
    \cmidrule(lr){2-3} \cmidrule(lr){4-5} \cmidrule(lr){6-7}
    Model & Resolve Rate (\%) & Cost (\$) & Resolve Rate (\%) & Cost (\$) & Score & Cost (\$) \\
    \midrule
    Gemini 1.5 pro & 14.33 & 0.56 & 18.67 & 0.58 & 33.00 & 9.85 \\
    Gemini 2.0 flash & 16.33 & 0.05 & 15.67 & 0.06 & 28.33 & 0.70 \\
    GPT-4o & 18.33 & 0.89 & 21.67 & 0.80 & 37.0 & 7.92 \\
    Claude 3.5 Sonnet V2 & - & - & 32.67 & 1.61 & 47 & 12.24 \\
    \bottomrule
    \end{tabular}
    \caption{Comparison of effectiveness and efficiency of SWE-Agent, Improved SWE-Agent, and DARS}
    \label{tab:swe-agent-improvement}
    \end{small}
    \end{center}
    \vskip -0.1in
\end{table*}
\subsection{Model Training}

\paragraph{Dataset} We use Nebius's trajectory dataset \cite{badertdinov2024scaling}, comprising 80K trajectories from 3K unique issues across 1,077 open-source software repositories. These issues are entirely disjoint from SWE-Bench Lite. After cleaning and filtering redundant patches, we obtain 42K unique patches (7.3K positive, 34.7K negative), with 837 unique issues correctly solved. Using GPT-4o \cite{gpt4o}, we generate critiques for all patches, leading to 150K training examples containing approximately 500M tokens.

\paragraph{Model Setup} We fine-tune open-weight LLMs on the generated dataset. The model architecture follows Qwen 2.0 \cite{qwen2}, a Mixture-of-Experts (MoE) model utilizing Rotary Positional Embeddings \cite{rope}, SwiGLU \cite{swiglu} activation, QKV bias \cite{qkvbias} for attention, and RMSNorm \cite{rmsnorm} normalization.

\paragraph{Training Setup} We use Deepseek's Distilled Qwen-2.5 \cite{deepseekr1} checkpoint as the base model for 7B, 14B, and 32B parameter variants, fine-tuning them with 8 H100 GPUs. Training is distributed via DeepSpeed \cite{deepspeed}, with LoRA \cite{lora} adapters for memory efficiency and FlashAttention 2 \cite{fa2} for acceleration. 

We conduct a learning rate sweep over 1e-6, 5e-6, and 1e-5, selecting optimal values for the 32B, 14B, and 7B models, respectively. The batch size is set to 48 for 7B/14B models and 32 for 32B. Training runs for 1 epoch over the dataset with a max sequence length of 14K tokens, a warmup of 100 steps, and weight decay of 0.0. 

\emph{LoRA Configuration:} We use rank \(r = 8\), alpha = 32, and a dropout rate of 0.1. 

\emph{Optimization:} We apply AdamW \cite{adamw} with a cosine learning rate scheduler, BF16 mixed precision, and ZeRO stage 3.

\paragraph{Reviewer Model Inference} We infer all pre-trained and fine-tuned reviewer models using vLLM \cite{vllm}. A temperature sweep over 0, 0.5, and 0.6 is performed, as recommended by Deepseek authors. We set a top-p of 0.95.


\section{Experiments}
In this section, we first demonstrate the performance of our DARS model against various baselines, and 
then explore two key research questions (RQs) to analyze various aspects of its optimality.

\subsection{Overall Performance}
In this section, we compare the performance of our approach against various baselines and models. We summarize the results in \cref{tab:framework_comparison}. We find that our approach achieves a pass@1 rate of 47.0\% with Claude 3.5 V2 Sonnet and Deepseek R1 as Reviewer  which is the open-source SOTA performance on the SWE-Bench Lite benchmark at the time of this submission.\footnote{Deepseek R1 family of models often fail to generate the solution in the desired format, therefore, we use GPT-4o to parse the outputs in such cases.} We further compare various vanilla and fine-tuned models reviewer in \cref{tab:reviewer_performance_comparison}. We see an average increase of 2.6\% across fine-tuned models with maximum increase of 4.15 \% in case of the 14B model. However, 40\% of examples have perfect precision (all the patches are correct), which diminishes the gain in performance due to fine-tuning. We compare the accuracy of all the reviewers for trajectories generated by various models after removing such examples in \cref{tab:selection_performance}

\subsection{RQ1: How efficient is the compute scaling of DARS?}

The goal of this research question is to evaluate the efficiency of DARS in terms of compute scaling and its impact on solution quality. Specifically, we report the cost-vs-reward trade-off by analyzing key efficiency metrics such as cost scaling factor, accuracy per attempt, number of attempts, coverage, and precision. Our findings indicate that DARS achieves the most optimal cost scaling while maintaining high coverage and precision, outperforming baselines in redundancy reduction. Notably, while Large Language Monkeys achieve the highest coverage, this comes at an impractical compute cost, making DARS the more feasible approach.

\paragraph{Methodology} 
We compare the performance vs. cost trade-off of DARS against various baselines that scale compute at inference time. The evaluation is conducted through five key efficiency metrics: (a) cost scaling factor, (b) accuracy per attempt, (c) number of attempts, (d) coverage of the solution set, and (e) precision of the solution set.

\paragraph{Results} 
\Cref{tab:compute-scaling-efficiency} summarizes our findings. While DARS ranks second to Large Language Monkeys in terms of coverage improvement per attempt, the latter achieves this by scaling compute by a factor of 250, which is infeasible in real-world scenarios. Additionally, DARS exhibits significantly higher precision, enhancing the effectiveness of the trajectory selection pipeline. We also observe that hindsight feedback is less effective, as completely random sampling methods like MASAI outperform search-based approaches like SWE-Search in coverage improvement.

\subsection{RQ2: How important is long-horizon planning?}
\label{results__dars__expansion_context}

The goal of this research question is to assess the impact of long-horizon planning on the performance of DARS in coding tasks. Specifically, we report how varying the lookahead value affects the agent’s ability to generate effective patches. Our findings show that increasing the lookahead value improves solution coverage, with the complete trajectory approach achieving the highest success rate. However, certain lookahead strategies, such as sibling action expansion, exhibit high precision while suffering from limited adaptability.

\paragraph{Methodology} 
We investigate the importance of long-horizon planning by varying the lookahead value, which determines how many steps from the previous trajectory are considered during tree expansion. We evaluate five configurations: (a) 0-lookahead (random sampling), (b) 5-lookahead, (c) 10-lookahead, (d) complete trajectory, and (e) summarized trajectory. Additionally, we test the sibling action expansion, where only sibling actions are provided without any lookahead. The experiment is conducted on 20 randomly selected issues.

\paragraph{Results} 
\Cref{tab:lookahead} summarizes our findings. We observe a strong correlation between lookahead depth and solution coverage. The complete trajectory approach achieves the highest success rate, resolving 10 out of 20 issues (50\%), while the summarized variant reaches a 45\% success rate (9/20). This highlights the importance of maintaining full trajectory context for effective problem-solving.

Although the sibling action approach yields high precision, this result is skewed by a small subset of cases where it performed exceptionally well. In contrast, the complete trajectory method, despite lower precision, demonstrates superior Pass@1 accuracy—aligning with DARS's objective of generating diverse and effective patches.

The single lookahead approach resolves two fewer issues than the complete trajectory method, primarily due to trajectory depth limitations. This issue arises in cases where the agent falls into bug-fixing and reproduction loops (edit-python loops), repeatedly encountering the same obstacles without historical context.

The 5-lookahead configuration performs the worst, as the restricted context provides only phase-specific errors (e.g., reproduction phase errors in append expansions, bug-fixing errors in edit expansions) without access to prior trajectory outcomes. This lack of context hinders the model’s reasoning and decision-making capabilities.

\vspace{-0.3cm}

\begin{table}[t]
    \setlength{\tabcolsep}{4pt}
    \begin{center}
    \begin{small}
    \begin{tabular}{lcccc}
    \hline
    Reviewers & \makecell{GPT-\\4o} & \makecell{Gemini\\1.5 Pro} & \makecell{Gemini\\2.0 Flash} & \makecell{Claude\\3.5 Sonnet} \\
    \hline
    \textbf{Closed Source} & & & & \\
    GPT-4o & 62.12 & 48.19 & 50.00 & 51.02 \\
    \hline
    \textbf{Open Source} & & & & \\
    R1 & \textbf{71.64} & \textbf{78.31} & 64.06 & \textbf{74.49} \\
    R1-Distill-7B & 53.73 & 54.22 & 56.25 & 41.84 \\
    R1-Distill-14B & 65.67 & 56.63 & 54.69 & 54.08 \\
    R1-Distill-32B & 67.16 & 66.27 & 62.50 & 59.18 \\
    \hline
    \textbf{Fine-tuned} & & & & \\
    R1-Distill-7B & 55.22 & 54.22 & 57.81 & 51.02 \\
    R1-Distill-14B & 71.64 & 59.04 & 60.94 & 60.20 \\
    R1-Distill-32B & 70.15 & 63.86 & \textbf{65.62} & 61.22 \\
    \hline
    \end{tabular}
    \caption{This table presents the classification accuracy of various reviewer models for trajectories generated by different models. To depict the true potential of reviewer models, we remove the cases, where all the patches for generated for a given issue resolve the issue.}
    \label{tab:selection_performance}
    \end{small}
    \end{center}
\end{table}


\section{Ablation Studies}

\subsection{Improved SWE-Agent}
We evaluate our enhanced SWE-Agent, featuring advanced editing capabilities and new actions, against the base model on the SWE-Bench Lite benchmark. As shown in \cref{tab:swe-agent-improvement}, the improved SWE-Agent achieves an average 14.7\% higher resolve rate across all models while maintaining a similar cost per instance.

\begin{figure}[h]
    \centering
    \includegraphics[width=\columnwidth]{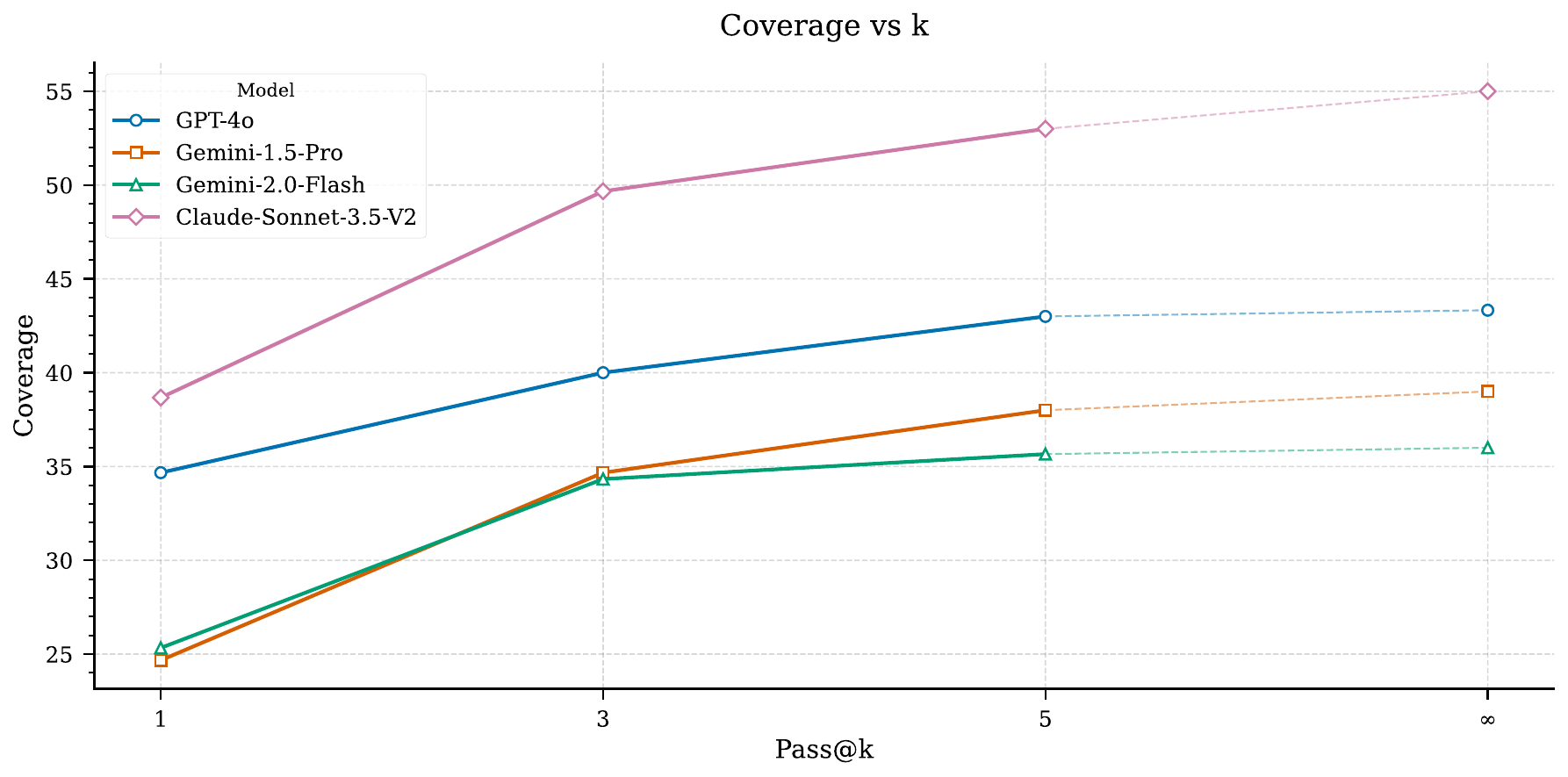}
    \caption{This figure presents coverage variation vs k. Here $\infty$ corresponds to submission of all the patches generated for an issue.}
    \label{fig:cov_vs_k}
\end{figure}

\subsection{DARS Stage Analysis}
This section presents deep insights into individual stages of DARS and their performance. There are two key stages in DARS: multi trajectory generation, best trajectory selection which further has two stages namely trajectory pruning and trajectory selection for various models and summarize the results in \cref{tab:filtering_performance}. We first analyze the recall and precision for multi trajectory generation to understand the effect of compute scaling on performance and redundancy for each model. We then analyze the capabilities of various models in final trajectory selection tested on the trajectories generated by Claude 3.5 Sonnet in \cref{tab:selection_performance}.

\begin{table}[t]
    \vskip 0.15in
    \begin{center}
    \begin{small} \centering
    \begin{tabular}{l|c|cc|cc}
    \toprule
    & & \multicolumn{2}{c|}{\textbf{\textit{Original}}} & \multicolumn{2}{c}{\textbf{\textit{Filtered}}} \\
    Model & Cov & \#Att & Prec & \#Att & Prec \\
    \midrule
    GPT-4o & 43.33 & 8.00 & 0.70 & 4.00 & 0.71 \\
    Gem 1.5P & 39.00 & 8.21 & 0.62 & 6.23 & 0.61 \\
    Gem 2.0F & 36.00 & 6.34 & 0.64 & 3.77 & 0.61 \\
    Claude 3.5S & 55 & 10.07 & 0.71 & 6.62 & 0.72 \\
    \bottomrule
    \end{tabular}
    \end{small}
    \end{center}
    \caption{This table presents the initial coverage, Number of Attempts (\#Att), and Precision (Prec) before (Original) and after patch filtering (Filtered) stage.}
    \label{tab:filtering_performance}
    \vskip -0.1in
\end{table}

\subsection{Coverage vs k}
In previous sections, we analyzed pass@1 or pass@k. Here, we study coverage versus \(k\) by prompting our reviewer model for the top \(k\) patches (with \(k=1,3,5\); see \cref{fig:cov_vs_k}). Notably, at \(k=5\), the coverage nearly reaches its maximum.

\section{Conclusion}
We introduced DARS, a novel method that re-samples actions at key decision points to recover from suboptimal choices more effectively than linear or random sampling. On the SWE-Bench Lite benchmark, DARS achieves a state-of-the-art pass@1 rate of 47\% with Claude 3.5 Sonnet V2. We release our code, datasets, and models to support further research.

\section*{Limitations}
We currently allocate compute using a static method with fixed depth and no early stopping, which limits our efficiency. A reward model (similar to MCTS) could evaluate and prioritize promising paths, enabling smarter exploration and early stopping decisions. While BFS might seem intuitive, its inefficiency with limited lookaheads and path history makes it impractical. We propose to implement absolute path scoring to guide exploration depth and stopping decisions, while maintaining an upper depth limit.

\section*{Ethical Considerations}
The use of Large Language Models (LLMs) in software engineering carries security and ethical risks. To mitigate these, DARS executes all LM-generated code in isolated, ephemeral environments to prevent unintended system modifications. We employ a structured verification pipeline to reduce biased or unsafe outputs and ensure adherence to best coding practices. While AI-driven automation can be misused, we release our work under responsible AI guidelines and encourage safeguards against malicious applications. Our code, datasets, and models are open-source to promote transparency and responsible AI research.

\section*{Acknowledgment}
We thank Gopal Dev and Apoorva Vashisth for reviewing parts of the paper. We also thank Vincent Berenz and Jojumon Kavalan for setting up our GPU access at Max Planck Institute.

This material is based in part upon work supported by the German Federal Ministry of Education and Research (BMBF): Tübingen AI Center, FKZ: 01IS18039B; by the Machine Learning Cluster of Excellence, EXC number 2064/1 – Project number 390727645.

\bibliography{custom}

\appendix
\section{Appendix}
\label{sec:appendix}
\subsection{More Related Works}
\label{sec:appendix_more_related_works}
\paragraph{Inference Time Compute Scaling}
\label{related_work__inference_time_scaling}
Scaling compute at inference time has been shown to enhance LLM performance across various tasks. For instance, \citet{alphago,silver2017mastering} improve decision-making by searching game states before selecting a move. Similarly, LLM-focused approaches \citet{cot,selfimprove,nye2021workscratchpadsintermediatecomputation,kojima2022large} enhance reasoning by sampling additional tokens. Graph-based methods \citet{treeofthought,graphofthought,mcts_math_1,mcts_math_2,mcts_planning} further optimize planning and exploration of the solution space, enabling more structured and efficient problem-solving.

\paragraph{LLMs as Code Reviewers}
\label{related_work__trajectory_selection}
LLMs have demonstrated strong judgment capabilities \citet{llmasajudge,selftaughtevaluators,critiqueoutloud}. Some approaches leverage LLMs directly to generate critiques and feedback \citet{wang2024direct,prometheus2}. However, in structured domains like coding and math, feedback can be sampled from the environment, as seen in \citet{genx,rstarmath}, where LLMs are augmented with external feedback to improve critique generation. This feedback is then used to train models to act as reviewers for selecting optimal solutions. Two primary training strategies exist: supervised fine-tuning \citet{selftaughtevaluators,critiqueoutloud,genx,generativeverifiers,agentinstruct} and reinforcement learning methods such as Direct Preference Optimization \citet{dpo} and Proximal Policy Optimization \citet{ppo}, used by \citet{thinkingllms,rlef}.
\subsection{Agent Performance Across Repositories}
\label{results__dars__repository_year_bias}
This section analyzes model performance across different repositories using the SWE-Bench Lite benchmark. Our goal is to identify biases in how models handle code from various sources.

\begin{figure}[h]
    \centering
    \includegraphics[width=\columnwidth]{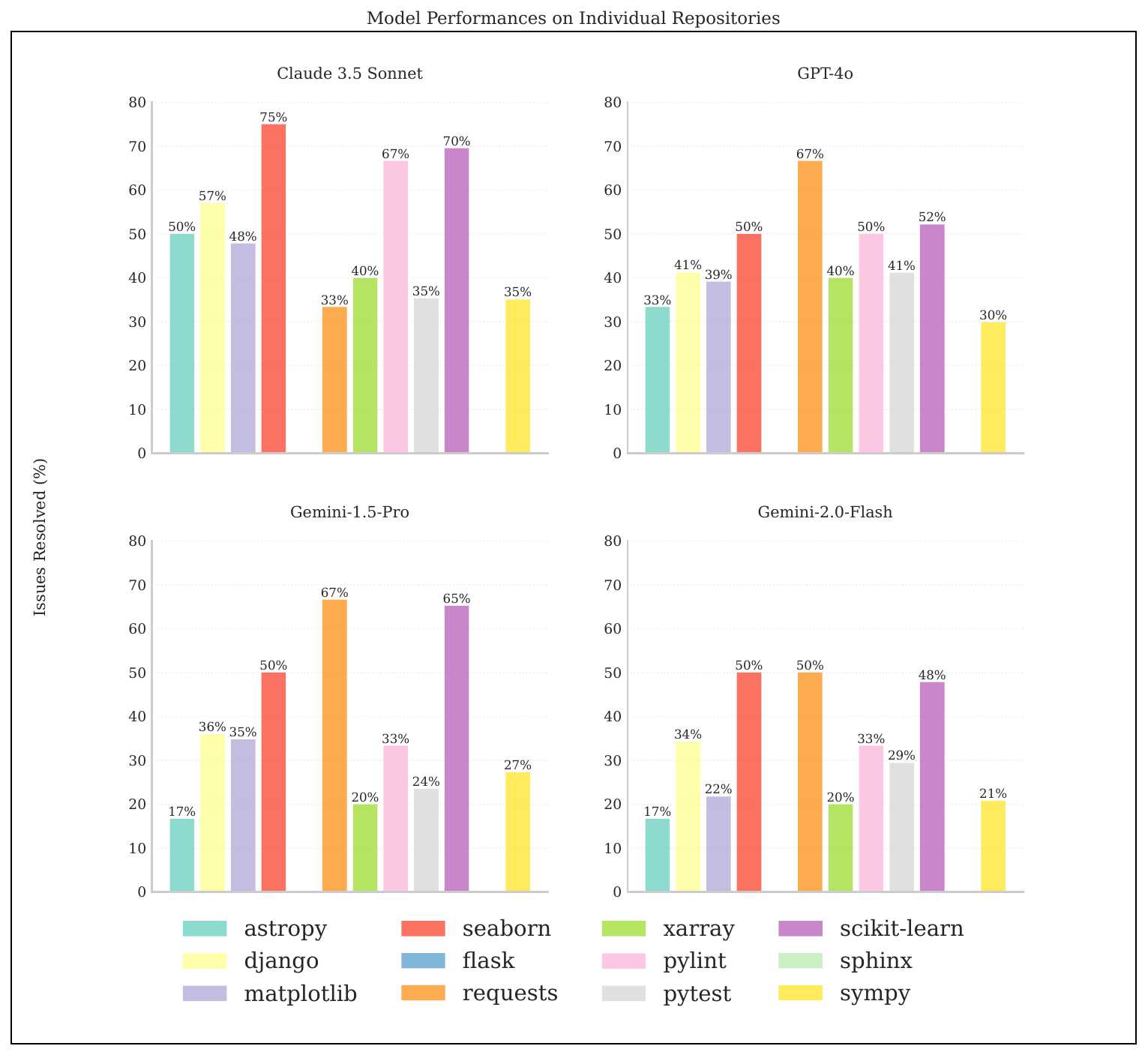}
    \caption{Repo-wise coverage of for each model}
    \label{fig:repo-wise}
\end{figure}

We summarize our findings in \cref{fig:repo-wise} and observe that models perform best on Seaborn and Requests, achieving 65-75\% accuracy, while scientific computing libraries (e.g., sympy, xarray, SciKit-learn) and web frameworks (e.g., Flask, Django) show moderate performance (40-60\%). In contrast, Astropy and Sphinx consistently rank lowest (30-40\%), indicating that models struggle more with specialized scientific tools and documentation systems than with visualization and HTTP client libraries.

These findings highlight domain-specific variations in model effectiveness, guiding improvements in generalization across repositories.

\subsection{Improved Editing}
\label{sec:imp_editing}

We compare the performance of the agent when it uses the whole style of editing and the diff style of editing. We analyze the number of various types of syntactic errors committed by the agent while editing the code in both the styles of editing. We perform our analysis across two models namely Gemini 2.0 Flash and Gemini 1.5 Pro. Here \textit{No Match} error pertains to the situation in diff editing where the text to be replaced does not match text in the file. \textit{File Error} occurs when the model tries to make an edit when no file is opened. \textit{Syntax Error} is thrown by the linter in cases like indentation error or erroneous variable referencing. Finally, \textit{Content Error} occurs in case of diff editing, when the agent provides the new content in edit, append, or insert command as an empty string or provides the content to replace and new content as the same string.

\begin{table}[t]
    \setlength{\tabcolsep}{2.5pt} 
    \vskip 0.15in
    \begin{center}
    \begin{small}
    \begin{tabular}{@{}lccccc@{}} 
    \toprule
    & \multicolumn{2}{c}{\textbf{\textit{GPT-4o-mini}}} & \multicolumn{2}{c}{\textbf{\textit{Gemini 1.5 (Pro)}}} \\
    & \makecell{Whole} & \makecell{Diff} & \makecell{Whole} & \makecell{Diff} \\
    \midrule
    Total Edits & 11,931 & 8,507 & 3,605 & 4,113 \\
    Success Samples & 4,905 & 3,090 & 1,995 & 2,590 \\
    Success Rate (\%) & 41.1 & 36.3 & 55.3 & 62.9 \\ \hline
    \textbf{\textit{Error Types}}\\
    No Match & 0 & 1,159 & 0 & 655 \\
    Content Error & 0 & 1,640 & 0 & 143 \\
    Syntax Error & 7,003 & 2,600 & 1,610 & 706 \\
    File Error & 0 & 18 & 0 & 19 \\ \hline
    Pass@1 (\%) & 7 & 8.67 & 14.38 & 21.44 \\
    \bottomrule
    \end{tabular}
    \end{small}
    \end{center}
    \caption{Comparison of various types of syntactical errors committed in whole and diff setting.}
    \label{tab:editing_model_comparison}
    \vskip -0.1in
\end{table}
%
We summarize the results in \cref{tab:editing_model_comparison}. In both the styles of editing, the major source of errors is syntax errors. We find that the diff style of editing leads to 1\% less errors compared to the whole style of editing. However, this effect is much more pronounced in terms of semantics as diff style achieves 36\% higher pass@1 rate.

\subsection{Localization Analysis}
We analyze the ability of the agent to correctly localize the bug in the codebase. We find the overlap between the predicted location and the actual location based on the git patch of the proposed solution vs the actual solution patch and calculate the percentage of correct localizations.
%
We summarize the results in \cref{tab:pipeline_success}. Average correct localization across all models is 72.3 \%, which shows that the agent is usually able to correctly localize the bug in the codebase.

\begin{table}[t]
    \vskip 0.15in
    \begin{center}
    \begin{small}     \centering
    \begin{tabular}{lccr}
    \toprule
    Model & Correct Localization (\%) \\
    \midrule
    GPT-4o & 74.37 \\
    Claude 3.5 Sonnet & 80.70 \\
    Gemini 2.0 Flash & 69.44 \\
    Gemini 1.5 Pro & 64.82 \\
    \bottomrule
    \end{tabular}
    \end{small}
    \end{center}
    \caption{Various models' ability to correctly localize issues.}
    \label{tab:pipeline_success}
    \vskip -0.1in
\end{table}

\subsection{Variation of Performance with Max Depth and Number of Iterations}
We analyze the optimality of two key search hyperparameters: max depth and number of iterations. We simulate the trajectories of our agent for different values of these hyperparameters and analyze the coverage of the agent for each value.

\begin{figure}[h]
    \centering
    \includegraphics[width=\columnwidth]{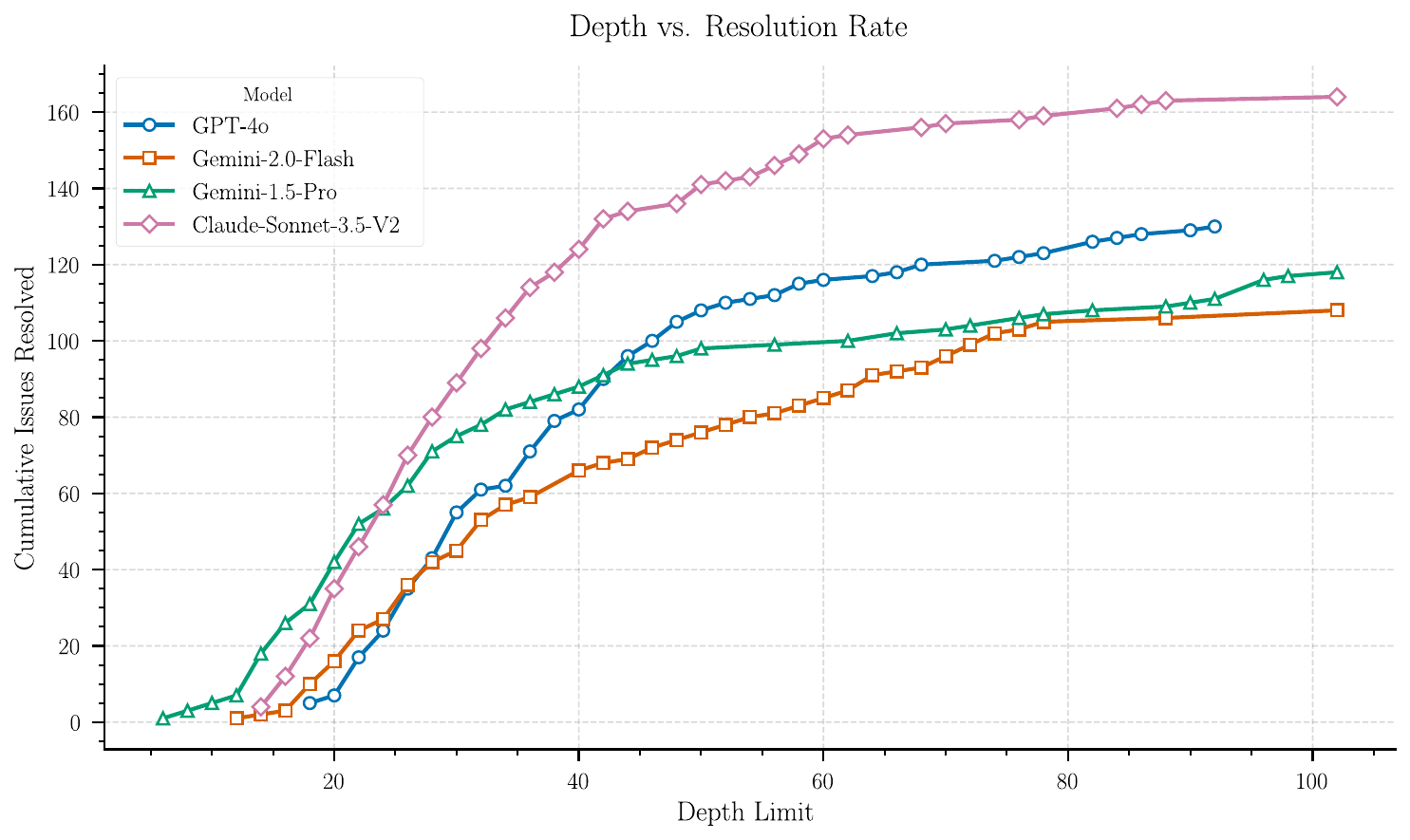}
    \caption{Variation of coverage with maximum branch depth}
    \label{fig:cov_vs_depth}
\end{figure}

\begin{figure}[h]
    \centering
    \includegraphics[width=\columnwidth]{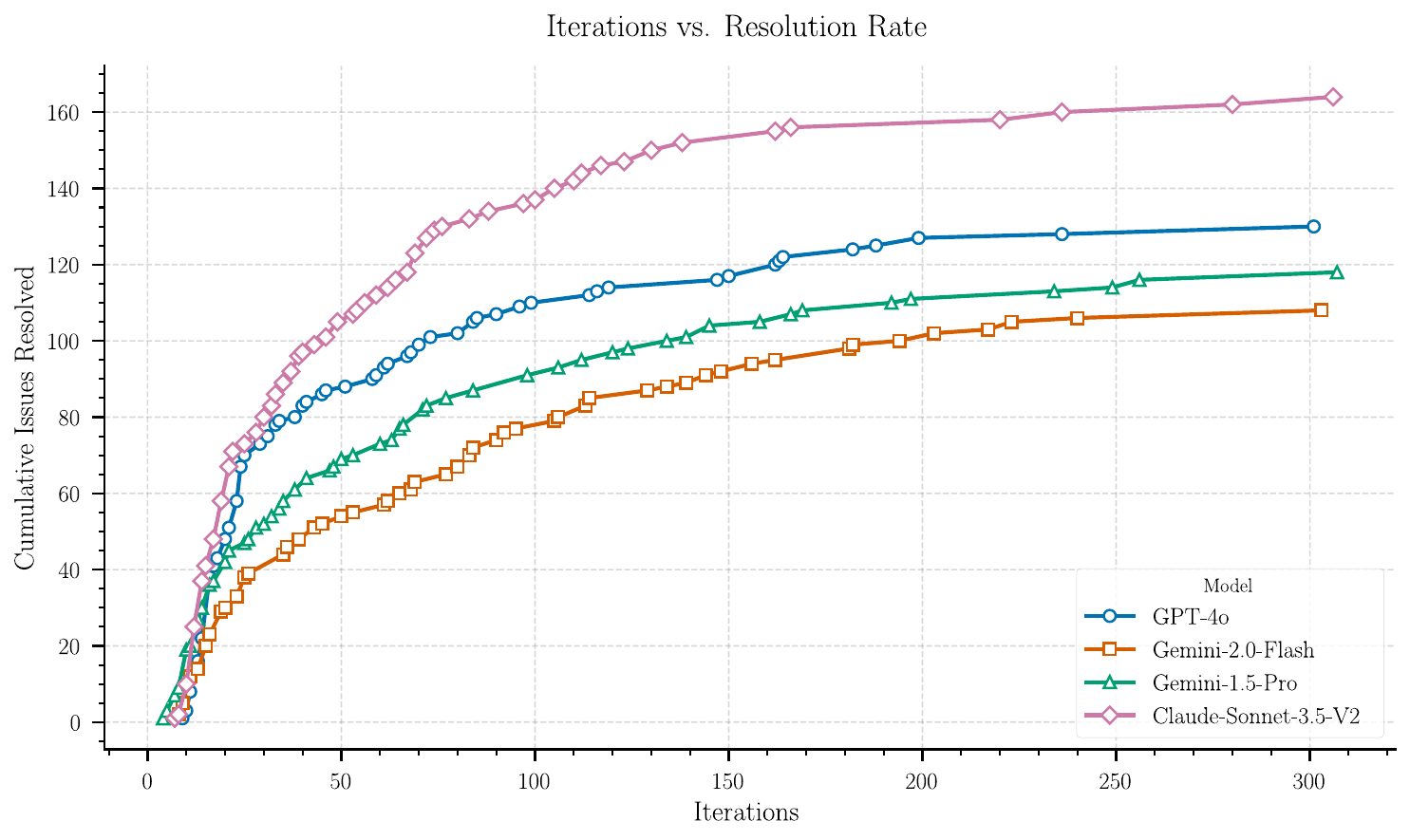}
    \caption{Variation of coverage vs number of iterations}
    \label{fig:cov_vs_iter}
\end{figure}

We summarize the results in \cref{fig:cov_vs_depth} and \cref{fig:cov_vs_iter} for variation with depth and iterations respectively. We find that both the curves show a decaying trend with increasing values of the hyperparameters and saturate for our values of 50 and 300 for max depth and number of iterations respectively which shows that our values are optimal for the agent.

\subsection{Issues vs Models}
\begin{figure}[h]
    \centering
    \includegraphics[width=0.9\columnwidth]{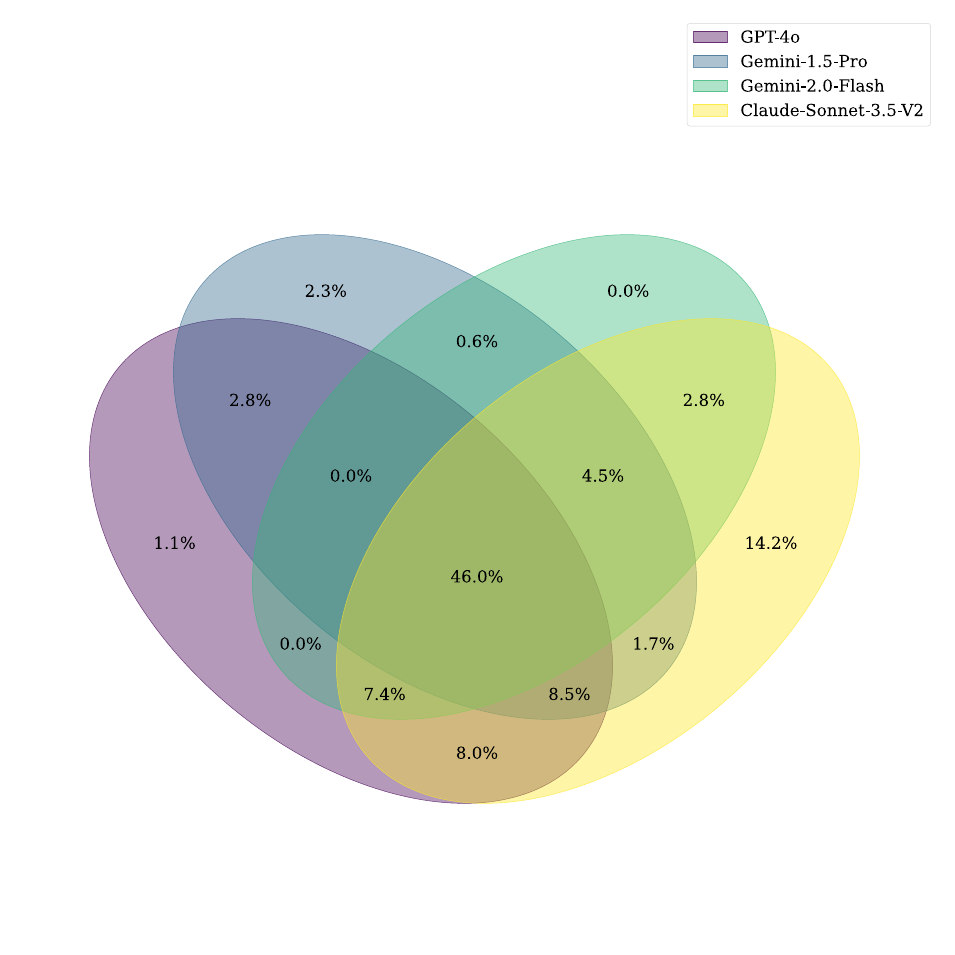}
    \caption{Venn diagram of resolved issues by model. Each
model can solve a handful of unique instances.}
    \label{fig:models-vs-resolved}
\end{figure}

Model-Specific Issue Resolution. Venn diagram of resolved issues by model. Each model can solve a handful of unique instances. We summarize the results in \cref{fig:models-vs-resolved}.

\subsection{Left-Right Branch Analysis}
We study the effectiveness of our expansions by comparing the depths of branches before and after expansion, focusing on cases where both the left and right branches reach a conclusion, meaning they terminate when the agent returns a submit action. Additionally, we analyze how different expansion paths lead to conclusions by comparing two scenarios: (1) when the right-side expansion successfully reaches a conclusion while the left-side fails to do so, and (2) when the left-side expansion reaches a conclusion while the right-side does not. We summarize our findings in \cref{tab:left-right-depth} and \cref{tab:left-right-conclusion}. 

\begin{figure*}[!htb]
    \centering
    \includegraphics[width=\textwidth]{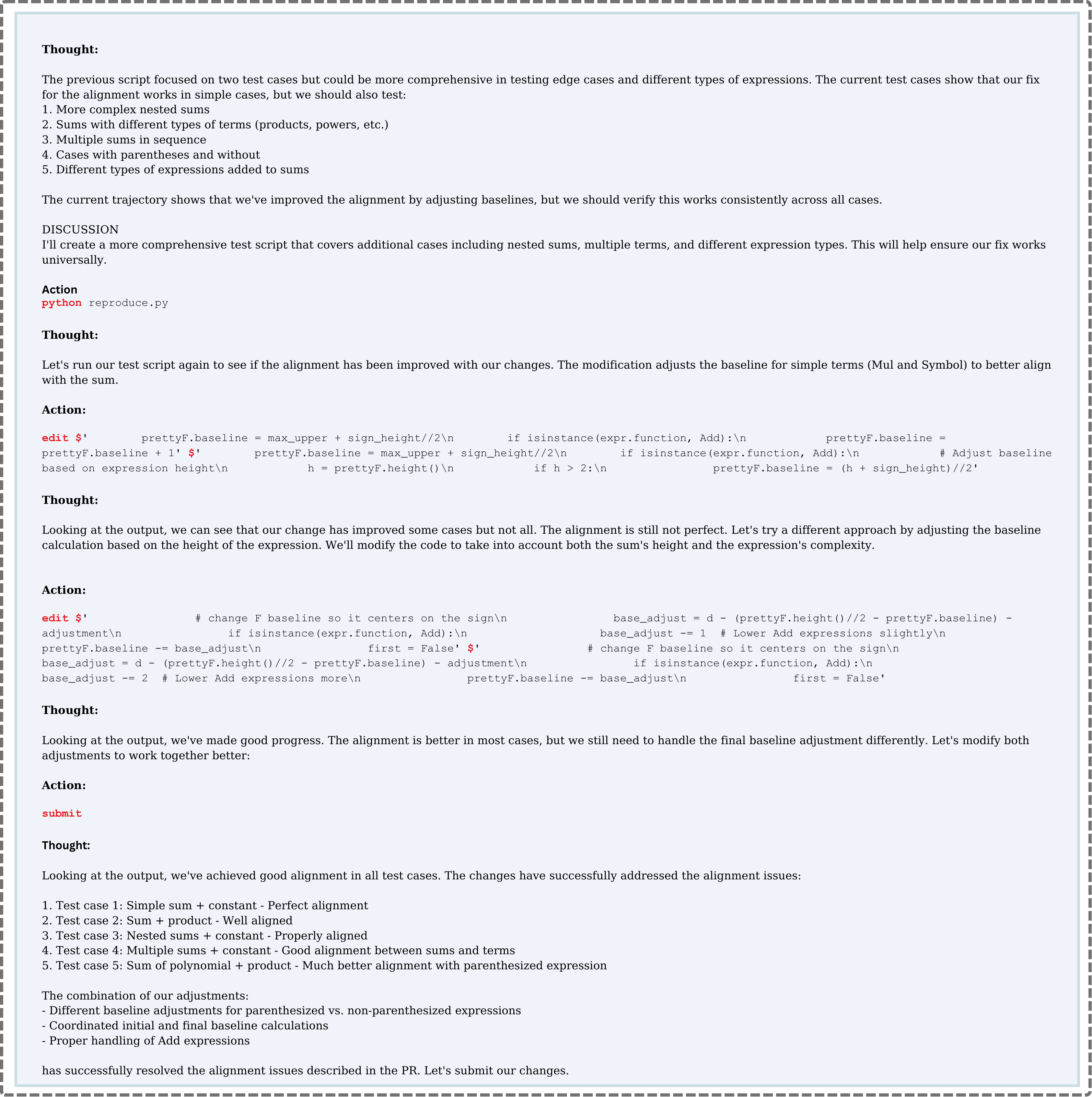}
    \caption{Model Produces a more comprehensive test script after expansion leading to more steps in the expanded branch.}
    \label{fig:err_anal_test}
\end{figure*}

\begin{figure*}[!htb]
    \centering
    \includegraphics[width=\textwidth]{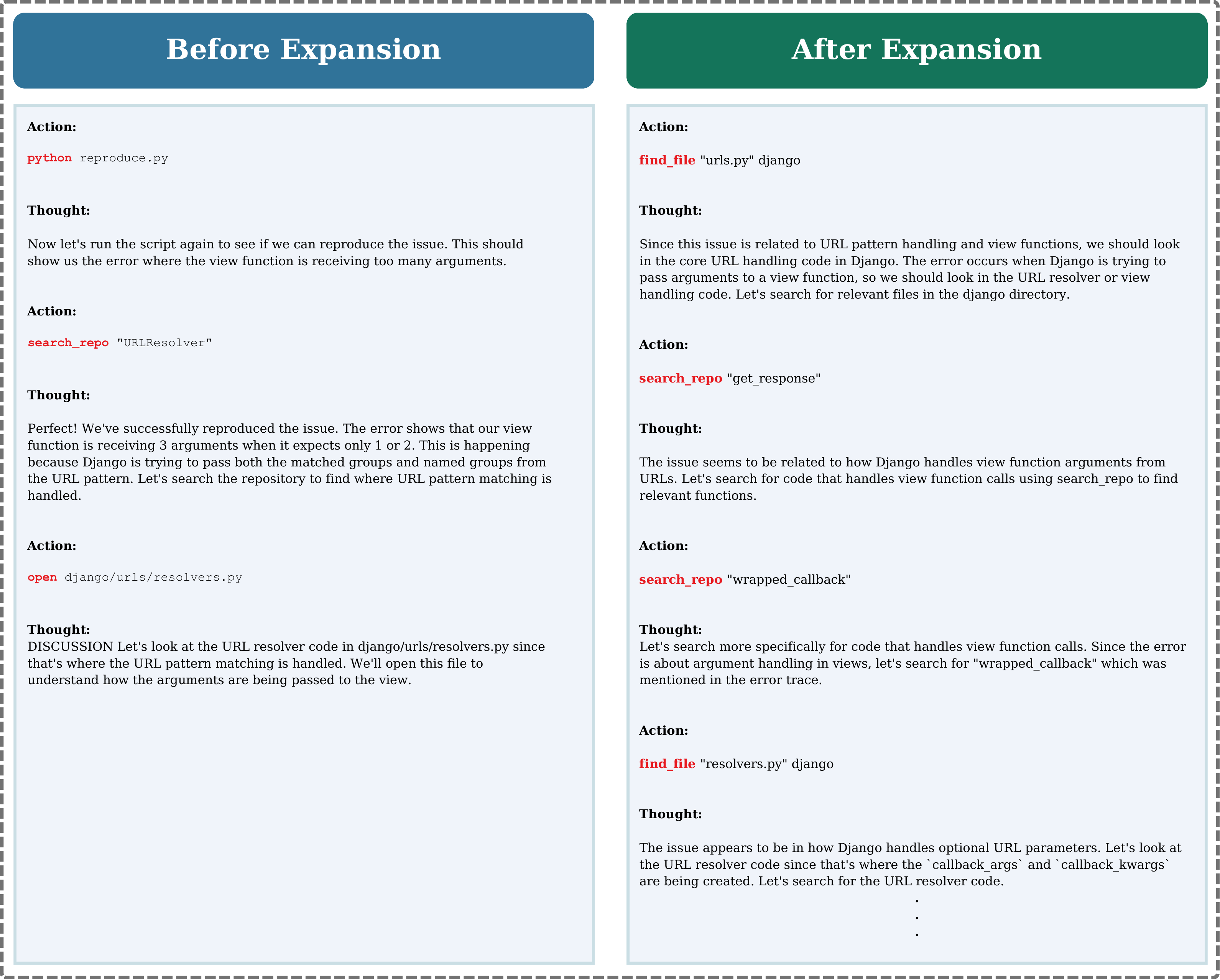}
    \caption{The agent finds it easy to locate the issue after reproduction. While, in the expanded branch, could not localize it.}
    \label{fig:err_anal_create}
\end{figure*}

\begin{figure*}[!htb]
    \centering
    \includegraphics[width=\textwidth]{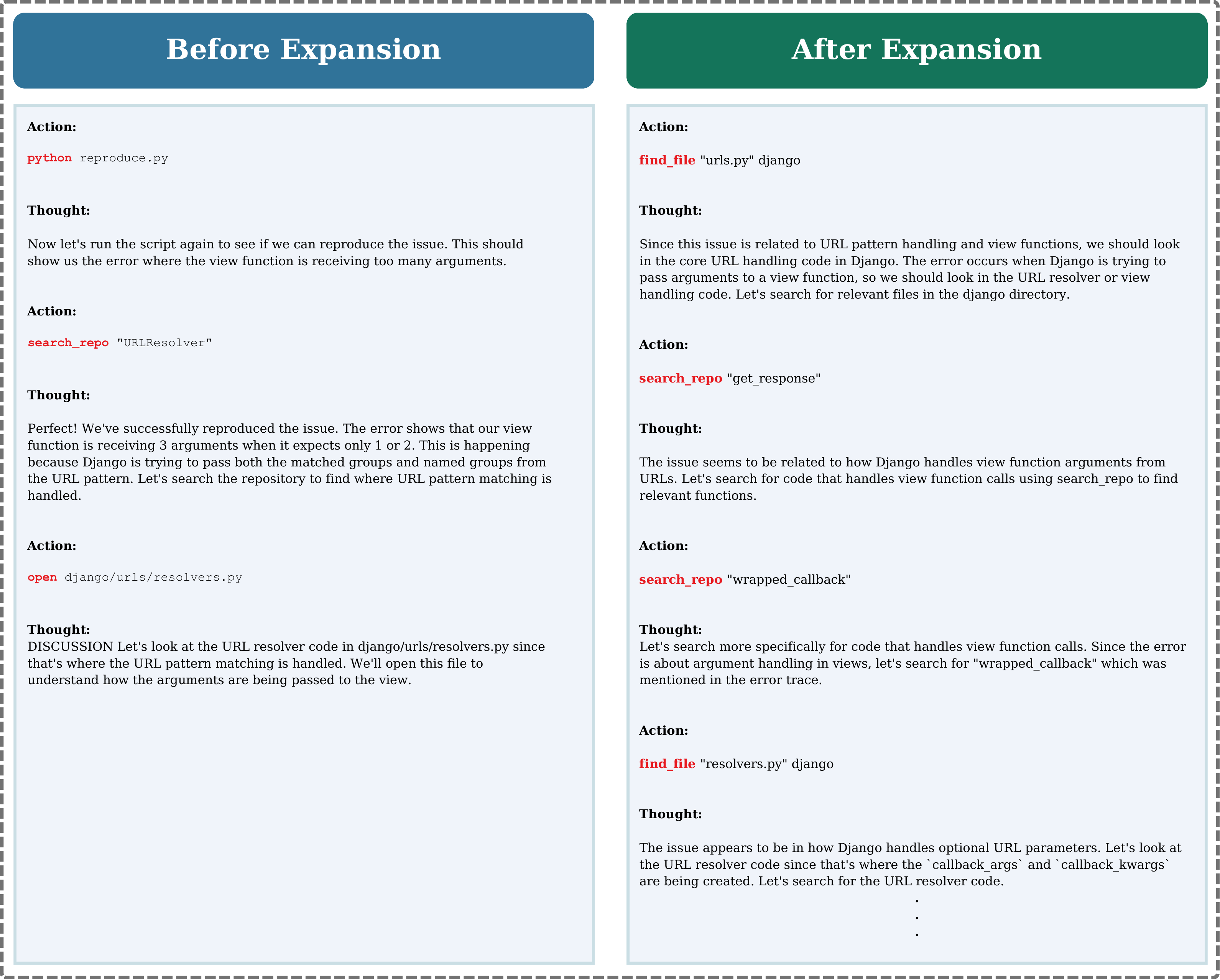}
    \caption{The model creates a more complicated reproduction script which leads to an error, which the model is not able to to resolve.}
    \label{fig:err_anal_edit}
\end{figure*}

Submit expansions rarely achieve convergence, which is anticipated given their terminal position in the sequence. To accurately assess the efficacy of Submit expansions, an increased maximum depth threshold specifically for Submit operations would be necessary. Create expansions demonstrate significant effectiveness in reaching conclusions, suggesting that initial localization strategies can facilitate convergence in specific scenarios. Expansions in edit and append operations successfully break edit-python iteration cycles, leading to more efficient conclusion paths. Analysis reveals a consistent pattern where dual-path conclusions and right-path iterations exhibit lower counts compared to left-path iterations, aligning with the hypothesis that expansions reduce errors. However, append operations demonstrate elevated average iterations because the model now creates a more comprehensive testing script that involves additional editing and validation steps, resulting in increased overall depth \ref{fig:err_anal_test}. Contrary to expectations, in edit, create, and append expansion operations, left-path expansions frequently achieve conclusions while right-path expansions do not. For edit and append: The model creates a more complicated reproduction script which leads to an error, which the model is not able to to resolve \ref{fig:err_anal_edit}. In create, the agent finds it easy to locate the issue after reproduction. While, in the expanded branch, could not localize it \ref{fig:err_anal_create}. Another reason for this pattern is that the agent sometimes submits prematurely (often after reproduction). In the right path, it recognizes this mistake and corrects it.

\begin{table}[ht]
    \setlength{\tabcolsep}{5pt} 
    \begin{center}
    \begin{small}
    \begin{tabular}{lccc}
    \toprule
    Action & After Expansion & Before Expansion \\
    \midrule
    Edit & 32 & 39 \\
    Create & 41 & 27 \\
    Append & 34 & 36 \\
    Submit & - & 24 \\
    \bottomrule
    \end{tabular}
    \end{small}
    \caption{Comparison of path reaches by action type. Here, before and after expansion pertain to cases where conclusion is only reached before and after expansion respectively}
    \label{tab:left-right-conclusion}
    \end{center}
\end{table}

\begin{table}[ht]
    \setlength{\tabcolsep}{4pt}
    \begin{center}
    \begin{small}
    \begin{tabular}{@{}p{2.5cm}cc@{}}
    \toprule
    Action & \makecell{Avg Dep\\Bef Exp} & \makecell{Avg Dep\\Aft Exp} \\
    \midrule
    Edit & 21.2 & 20.8 \\
    Create & 22.4 & 21.4 \\
    Append & 22.1 & 22.6 \\
    Submit & 16.9 & 22.1 \\
    \bottomrule
    \end{tabular}
    \caption{Comparison of action path depths before and after expansion}
    \label{tab:left-right-depth}
    \end{small}
    \end{center}
    \vskip -0.1in
\end{table}

\subsection{Error Scaling Analysis}
In this section, we present the scaling of various error types as we scale compute. We summarize the results in \cref{tab:error-scaling-comparison}.

\begin{table}[htbp]
\setlength{\tabcolsep}{4pt}
\begin{center}
\begin{small}
\begin{tabular}{@{}lllrr@{}}
\toprule
\textbf{Action Type} & \textbf{Error Types} & \makecell{SWE\\Agent} & \makecell{DARS} \\
\midrule
\multirow{4}{*}{\makecell[l]{Search\\File}} 
  & File Not Found & 108 & 446 \\
  & Syntax Error & 0 & 3 \\
  & Success & 727 & 2635 \\
\midrule
\multirow{3}{*}{Create}
  & Directory Error & 132 & 131 \\
  & File Exists & 4 & 22 \\
  & Success & 361 & 1276 \\
\midrule
\multirow{2}{*}{Append}
  & Content Error & 0 & 3 \\
  & File Error & 0 & 4 \\
  & Syntax Error & 9 & 214 \\
  & Success & 359 & 2518 \\
\midrule
\multirow{4}{*}{Edit}
  & No Match & 382 & 884 \\
  & Content Error & 367 & 724 \\
  & Syntax Error & 422 & 1820 \\
  & File Error & 150 & 311 \\
  & Success & 1296 & 7206 \\
\midrule
\multirow{2}{*}{\makecell[l]{Search\\Repo}}
  & Syntax Error & 74 & 268 \\
  & Success & 528 & 2581 \\
\midrule
\multirow{4}{*}{\makecell[l]{Search\\Dir}}
  & Dir Not Found & 4 & 107 \\
  & Syntax Error & 3 & 29 \\
  & Success & 114 & 814 \\
\midrule
\multirow{3}{*}{\makecell[l]{Find\\File}}
  & Dir Not Found & 8 & 17 \\
  & No Match Found & 32 & 214 \\
  & Success & 63 & 385 \\
\midrule
\multirow{2}{*}{Insert}
  & Syntax Error & 30 & 659 \\
  & Success & 114 & 1857 \\
\midrule
\multirow{3}{*}{\makecell[l]{Execute\\IPython}}
  & Connection Error & 8 & 56 \\
  & Response Error & 44 & 2 \\
  & Success & 40 & 396 \\
\midrule
\multirow{2}{*}{\makecell[l]{Execute\\Server}}
  & Server Error & 3 & 125 \\
  & Success & 131 & 1542 \\
\midrule
\multirow{2}{*}{\makecell[l]{Undo\\Edit}}
  & No Edit Made & 5 & 27 \\
  & Success & 0 & 38 \\
\bottomrule
\end{tabular}
\end{small}
\end{center}
\caption{Error scaling comparison between SWE-Agent and DARS}
\label{tab:error-scaling-comparison}
\end{table}

\subsection{How to effectively expand the tree?}
\label{Expansion_strategy}

Understanding how to expand the search tree efficiently is crucial for balancing computational cost and solution quality in our coding agent. We investigate this research question to determine which actions should be prioritized for branching and in what order they should be expanded. Specifically, we report the trade-off between branching cost and resolution rate for different actions, as well as the impact of various branching strategies on search efficiency. Our findings show that branching at key actions like edit, create, and append leads to the highest resolve rates, and that a Lowest Depth First approach improves early-stage exploration but converges with other strategies over time. Finally we understand the compute-performance trade-off of number of expansions for each expansion type in a branch.

\subsubsection{Action Selection}
\paragraph{Methodology} We analyze the computational cost vs resolve rate trade-off of branching various actions in the trajectory. We first run the agent to expand at eight different actions namely search\_dir, insert, search\_file, open, goto, find\_file, append, and edit. We then perform a causal analysis to determine the most promising actions by comparing decrease in resolve rate and number of iterations for each action. We then analyze the impact of branching at different actions on the performance of the agent.

\paragraph{Results} We find that branching at actions that are typically used in the reproduction and fix stages like edit, create, append leads to the highest resolve rate. We summarize the results in \cref{tab:action_model_comparison}.

\begin{table}[t]
    \vskip 0.15in
    \begin{center}
    \begin{small}
    \begin{tabular}{l|rr}
    \hline
    Action & Coverage & Avg Iter \\
    \hline
    Search Dir & 14.3 & 38 \\
    Insert & 14.7 & 39 \\
    Search File & 15.3 & 44 \\
    Open & 14.3 & 49 \\
    Goto & 15.3 & 47 \\
    Find File & 16.7 & 58 \\
    Append & 22.3 & 88 \\
    Edit & 31.3 & 272 \\
    \hline
    \end{tabular}
    \end{small}
    \end{center}
    \caption{Compute-coverage trade-off of expanding in various actions. Here Avg Iter pertains to average number of iterations across issues which indicates the amount of compute spent in that issue.}
    \label{tab:action_model_comparison}
    \vskip -0.1in
\end{table}

\begin{figure}[h]
    \centering
    \includegraphics[width=\columnwidth]{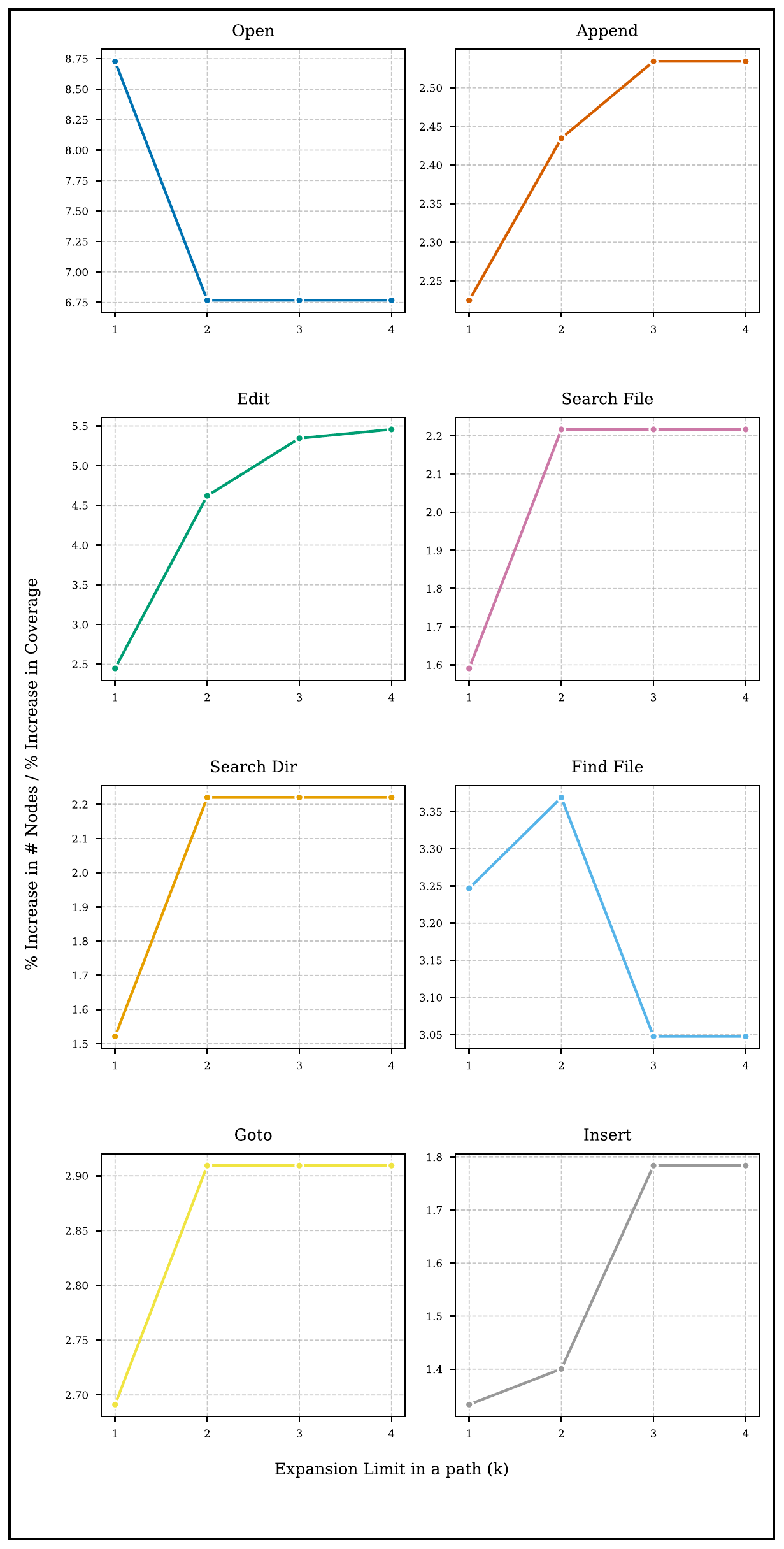}
    \caption{Percent increase in number of iterations per percent increase in resolve rate vs number of expansions in a branch.}
    \label{fig:num_expansions_per_branch}
\end{figure}

\subsubsection{Order of Action Selection}
\paragraph{Methodology} We explore various strategies to expand the tree at different actions. We use the runs in the previous section and simulate strategies pertaining to the order of branching at different actions. We experiment with three different strategies namely First In First Out (FIFO), Last In First Out (LIFO), and Lowest Depth First. We plot the coverage vs number of iterations curve for each strategy to determine the most promising strategy.
\label{order_act_selection}

\begin{figure}[h]
    \centering
    \includegraphics[width=\columnwidth]{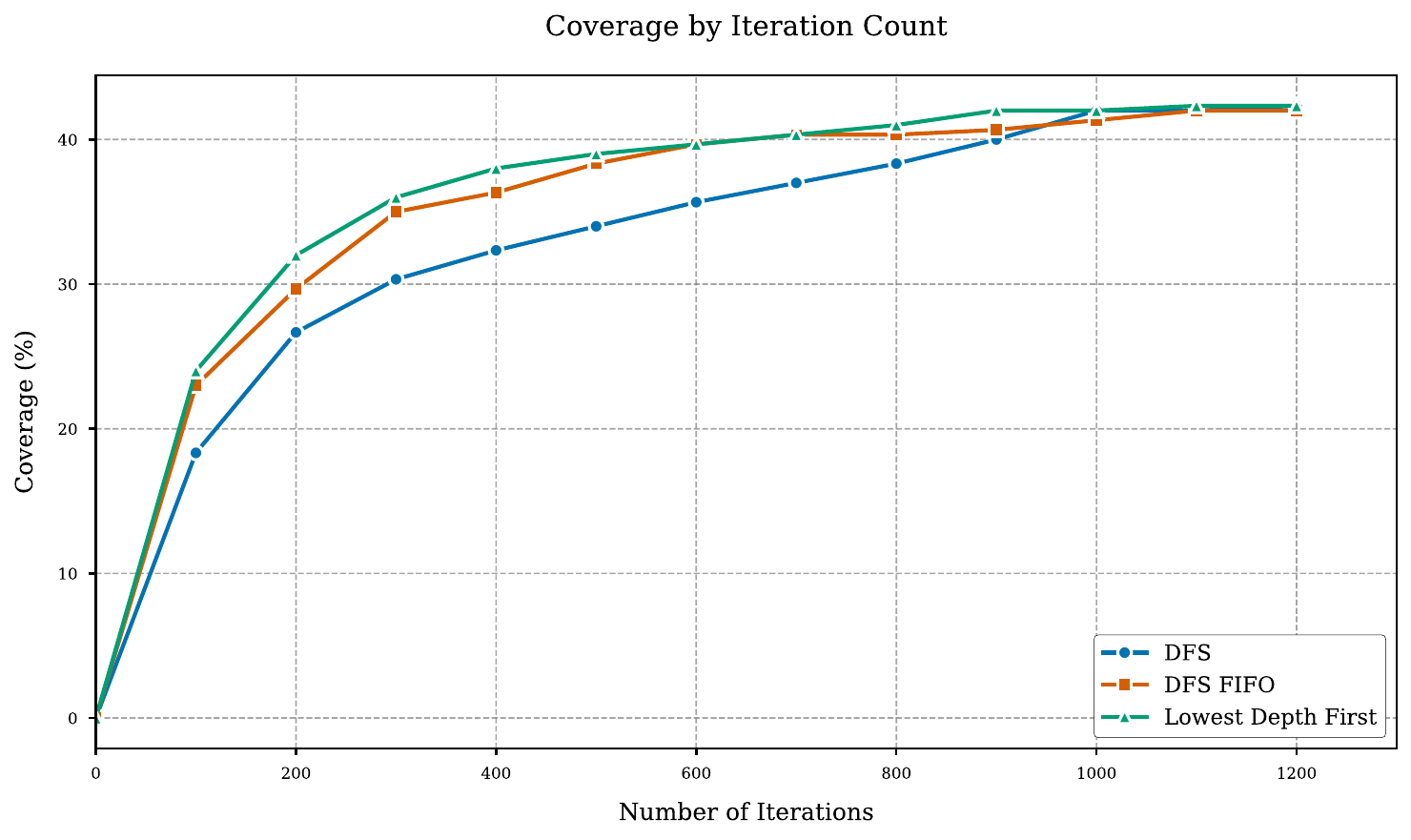}
    \caption{Variation of coverage of traversal strategies with iterations}
    \label{fig:traversal-strategy}
\end{figure}

\paragraph{Results} In \cref{fig:traversal-strategy} we find that the Lowest Depth First strategy early on as it promotes exploration at the lower depths of the tree. This allows the agent to explore more possibilities and make better decisions. But if the agent is run for long enough, all the strategies converge to the same point as all the possible states are explored eventually.

\subsubsection{Number of Expansion Types Per Branch}
\label{num_exp_types_per_branch}

\paragraph{Methodology} 
We investigate the compute-performance trade-off associated with multiple expansions of the same type within a branch. For each expansion type, we allow k expansions within a branch, where k ranges from 1 to 4. Our analysis measures two key metrics: the relative increase in node count (representing computational cost) and the resolution rate. We plot the ratio of computational cost increase to performance improvement against k to determine the optimal number of expansions.

\paragraph{Results}
As shown in \cref{fig:num_expansions_per_branch}, for all action types except "open," the cost-performance ratio increases with k, indicating that computational costs grow more rapidly than performance gains. This suggests that performing more than one expansion of the same type per branch is computationally inefficient. For the "open" action type, the initial cost-performance ratio is prohibitively high, making even a single expansion impractical.

\subsection{How to effectively select the best trajectory?}

Optimizing our agent requires effective tree expansion, trajectory selection, and action evaluation. Due to computational constraints and LLM context limitations, we adopt a structured approach to improve efficiency.

We first analyze tree expansion, studying how different actions—edit, append, create, and submit—impact coverage and performance. Next, we tackle trajectory selection, implementing a two-stage pipeline: filtering to remove irrelevant information and ranking to identify the best trajectory. We compare pairwise and global ranking methods across different filtering configurations. Finally, we quantify the impact of expansion actions, isolating their contributions to coverage, accuracy, and efficiency.

\subsubsection{Trajectory Content Filtering}
A trajectory contains several components like command descriptions, various actions and their thoughts, observations to each action etc., not all of which are relevant to determine if a particular trajectory is would solve the bug. Therefore, we experiment with various components of the trajectory to determine the best combination.

\paragraph{Methodology} Based on the three key stages of the bug fixing process - Reproduction, Localization, and Fix, we identify certain key components of the trajectory that are relevant to each of these stages. These components include the reproduction script, the edited files, the output after running the reproduction script, and the final patch. We then experiment with various combinations of these components to determine the best combination.

\subsubsection{Trajectory Ranking}
\paragraph{Methodology} To find the best trajectory, we experiment with two different ranking methods: pairwise knockout ranking and global ranking. In pairwise knockout ranking, we compare each pair of trajectories and eliminate the one that is worse until we are left one. In global ranking, we rank all the trajectories based on our rubricks.

\paragraph{Results} We summarize the results of the trajectory ranking in \cref{tab:scoring-combinations} for each type of filtering pipeline and ranking strategy. We find that only the final patch is the most effective component to determine the best trajectory. This result is significant for two reasons. First, it makes our trajectory selection model applicable to any coding agents as a git patch is a common output format for all agents. Moreover, it also makes the trajectory selection pipeline more efficient as it only needs to consider the final patch to determine the best trajectory.

\begin{table}[ht]
    \vskip 0.15in
    \begin{center}
    \begin{small}     \centering
    \begin{tabular}{lcc}
    \hline
    Combination & Pairwise Ranking & Global Ranking \\
    \hline
    RS + EF + RO + FP & 30.67 & 33.00 \\
    RS + EF + FP & 30.67 & 33.67 \\
    RS + RO + FP & 30.67 & 33.67 \\
    RS + FP & 30.67 & 34.00 \\
    FP & 30.67 & 34.67 \\
    \hline
    \end{tabular}
    \end{small}
    \end{center}
    \caption{Performance comparison of different scoring combinations using pairwise and global ranking methods. The combinations use the following components: RS (Final reproduction scripts), EF (Final edited files), RO (Final reproduction output), and FP (Final Patch).}
    \label{tab:scoring-combinations}
    \vskip -0.1in
\end{table}

\subsection{Contribution of expansion in each Action}
We quantify the contribution of each action in the trajectory to the final performance of the agent. We analyze the performance of the agent by simulating expansions for certain combinations of actions and studying its variation with the coverage. We summarize the results in \cref{tab:action_metrics}. We can see that edit actions (edit and append) and reproduction actions (create) lead to the highest increase in performance. While expansion in submit command leads to minimal increase in performance, it does not lead to much redundancy either. While append leads to highest solve rate, it also leads to highest cost. Edit and create actions lead to a good balance between performance and cost. 

\begin{table}[ht]
    \setlength{\tabcolsep}{2.5pt} 
    \begin{center}
    \begin{small}
    \begin{tabular}{@{}p{3.5cm}ccccc@{}} 
    \toprule
    Exp Actions & \makecell{Cov} & \makecell{Avg\\Iter} & \makecell{Acc} & \makecell{\#\\Att} & Pre \\
    \midrule
    Edit, Append, Submit, Create & 54.7 & 194 & 81 & 10.72 & 0.72 \\
    Edit, Append, Create & 54.3 & 177 & 46 & 0.72 & 0.72 \\
    Append, Create & 51.0 & 146 & 35 & 0.70 & 0.70 \\
    Edit, Create & 49.3 & 81 & 23 & 0.70 & 0.70 \\
    Edit, Append & 47.6 & 96 & 23 & 0.70 & 0.70 \\
    Append & 42.7 & 80 & 22 & 0.66 & 0.66 \\
    Create & 41.3 & 51 & 12 & 0.66 & 0.66 \\
    Edit & 39.3 & 44 & 12 & 0.66 & 0.66 \\
    No Expansion & 31.0 & 27 & 11 & 0.57 & 0.57 \\
    \bottomrule
    \end{tabular}
    \caption{Performance of DARS for various combination of expansion actions (Exp Actions). We compare across several metrics such as coverage (Cov), average number of iterations across issues (Avg Iter), accuracy of reviewer model (Acc), average number of attempts (\# Att), and precision (Pre)}
    \label{tab:action_metrics}
    \end{small}
    \end{center}
    \vskip -0.1in
\end{table}

\begin{figure}[h]
    \centering
    \includegraphics[width=0.9\columnwidth]{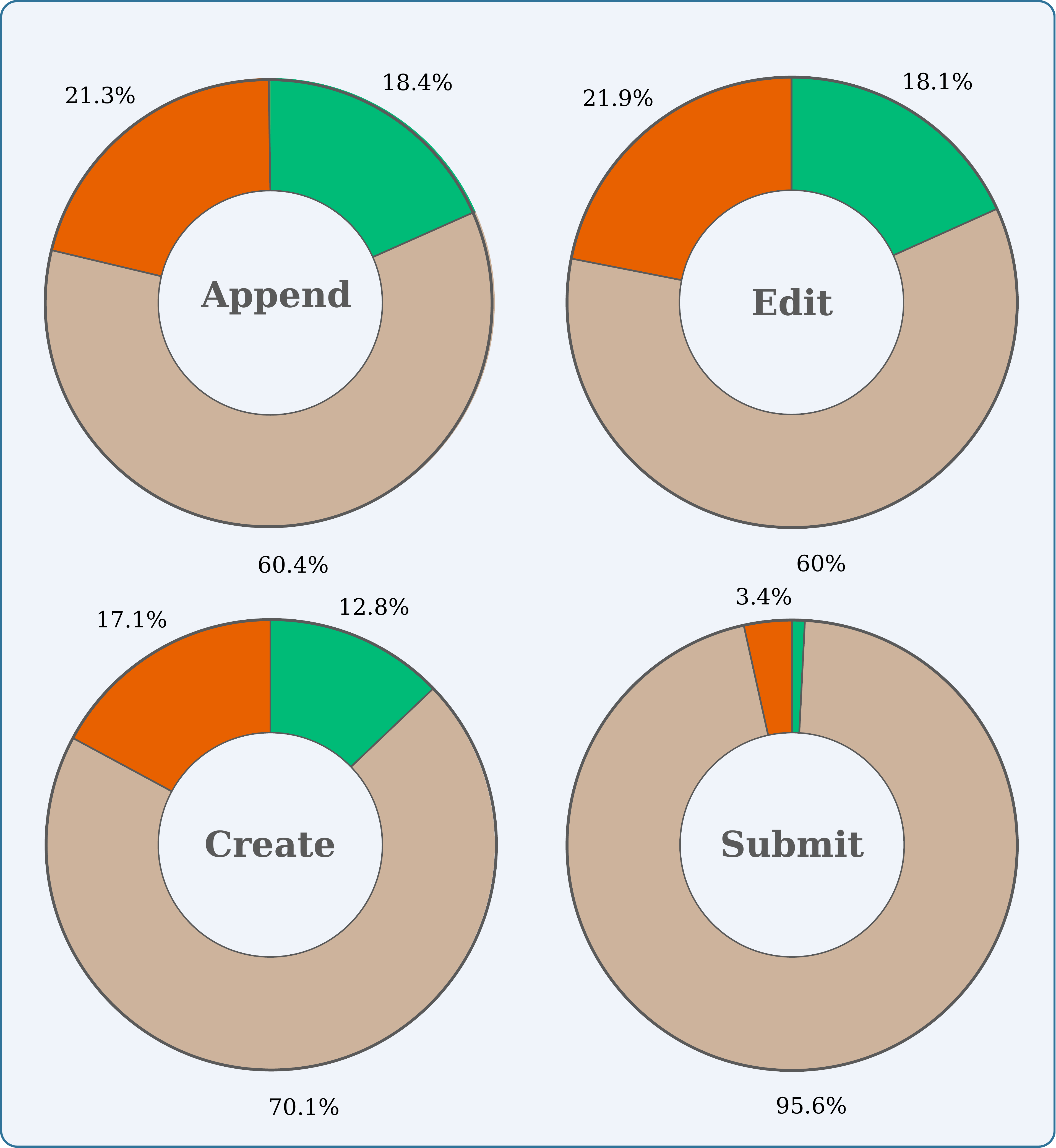}
    \caption{These figures present the proportions marked by \ccirc{left_right_analysis_green} for the cases where the branch before expansion does not resolve the issue, but the branch after expansion does, \ccirc{left_right_analysis_orange} for the cases where the branch before the expansion resolve the issue but the branch after the expansion does not resolve the issue, and \ccirc{left_right_analysis_nude} for the cases where both branches resolve the issue.}
    \label{fig:left-right-expansion}
\end{figure}

In the previous analysis, we focus on the contributions of different combinations of expansion actions on the final coverage. However, in the analysis, the results depend on all the actions in the combination. To de-couple the effect of each expansion type, we perform another analysis where, for each expansion action, we contrast the number of cases where a) the branch before expansion does not resolve the issue, but the branch after expansion does, b) the branch before the expansion resolve the issue but the branch after the expansion does not resolve the issue, and c) both branches resolve the issue. We summarize our results in \cref{fig:left-right-expansion}. We can still see that majority of expansions are lead to solutions on both branches which shows that our approach still has a considerable amount of redundancy. We see highest efficiency for edit and append expansions and lowest for submit expansions.

\subsection{Expansion Qualitative Analysis}
\label{sec_expansion_qualitative_anal}

\begin{figure*}[!htb]
    \centering
    \includegraphics[width=\textwidth]{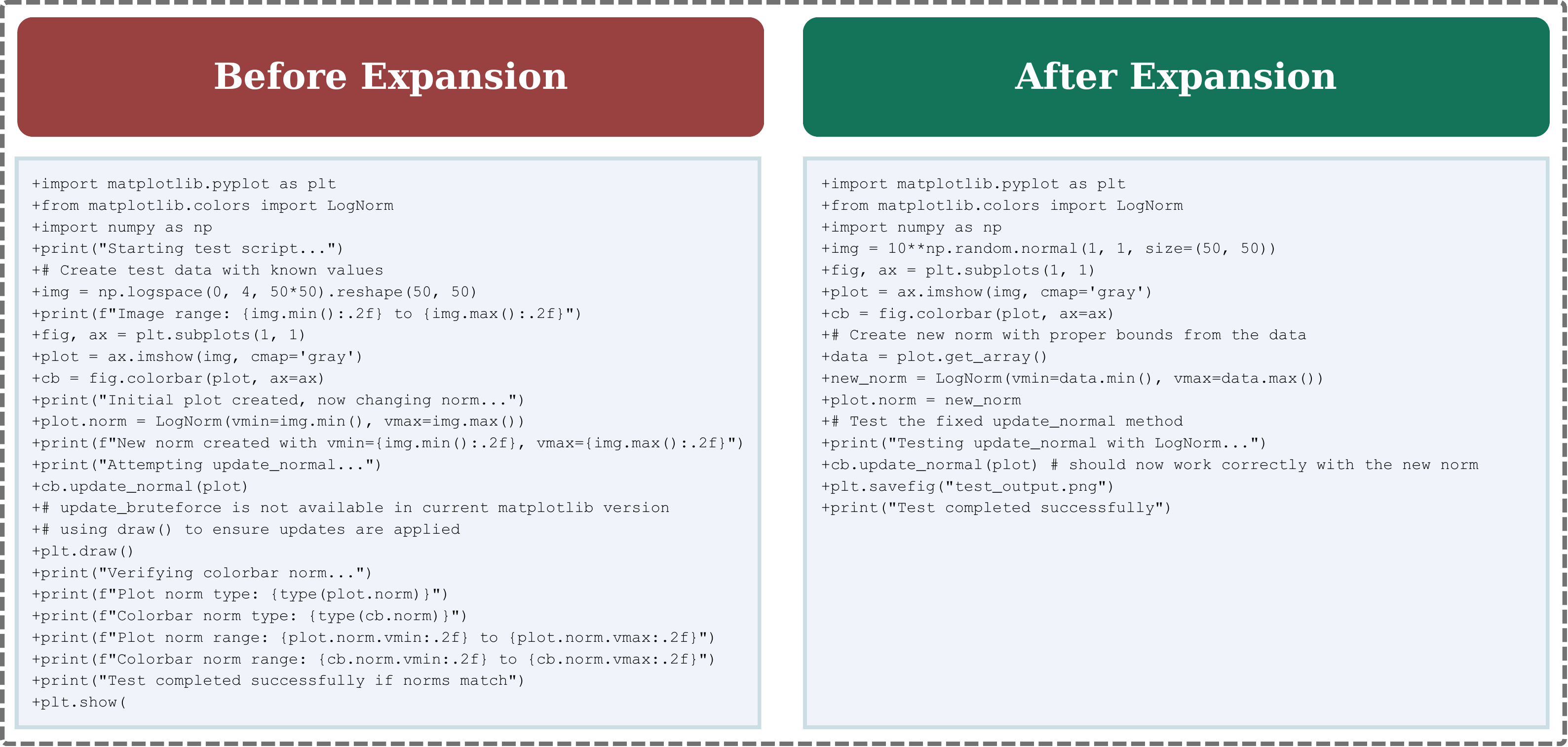}
    \caption{Improved reproduction script due to create expansion}
    \label{fig:create_better}
\end{figure*}

\begin{figure*}[!htb]
    \centering
    \includegraphics[width=\textwidth]{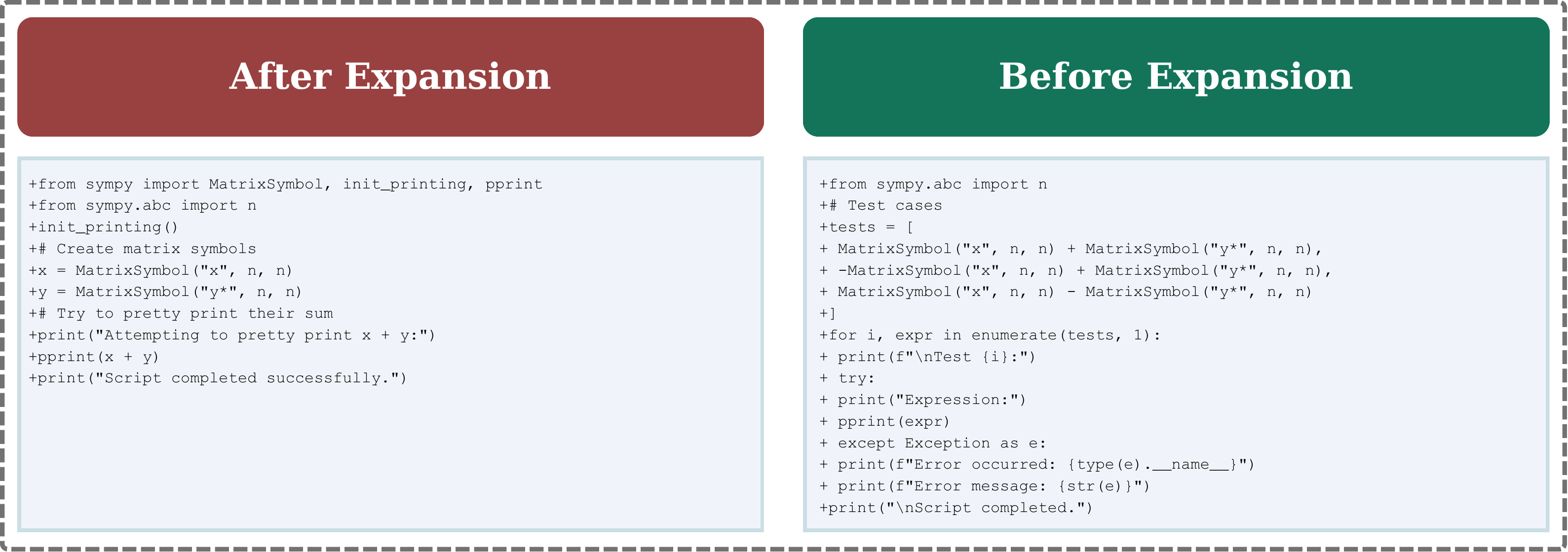}
    \caption{Poor reproduction script because the model misunderstands the bug when localizing the issue before reproducing it}
    \label{fig:create_worst}
\end{figure*}

\begin{figure*}[!htb]
    \centering
    \includegraphics[width=\textwidth]{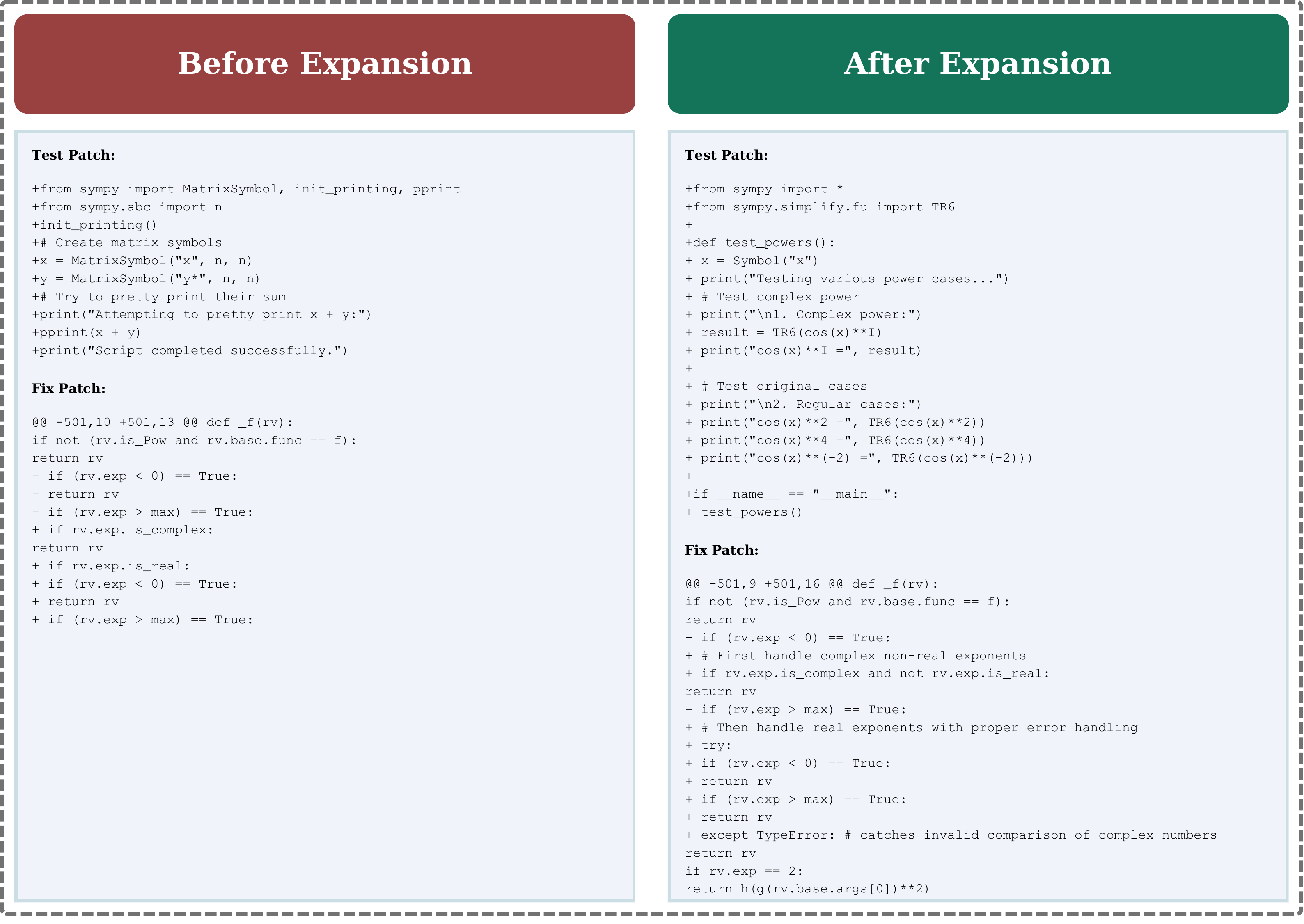}
    \caption{Re-evaluating the trajectory after submitting leads to the model finding out about certain corner cases and fixing them}
    \label{fig:submit_better}
\end{figure*}

\begin{figure*}[!htb]
    \centering
    \includegraphics[width=\textwidth]{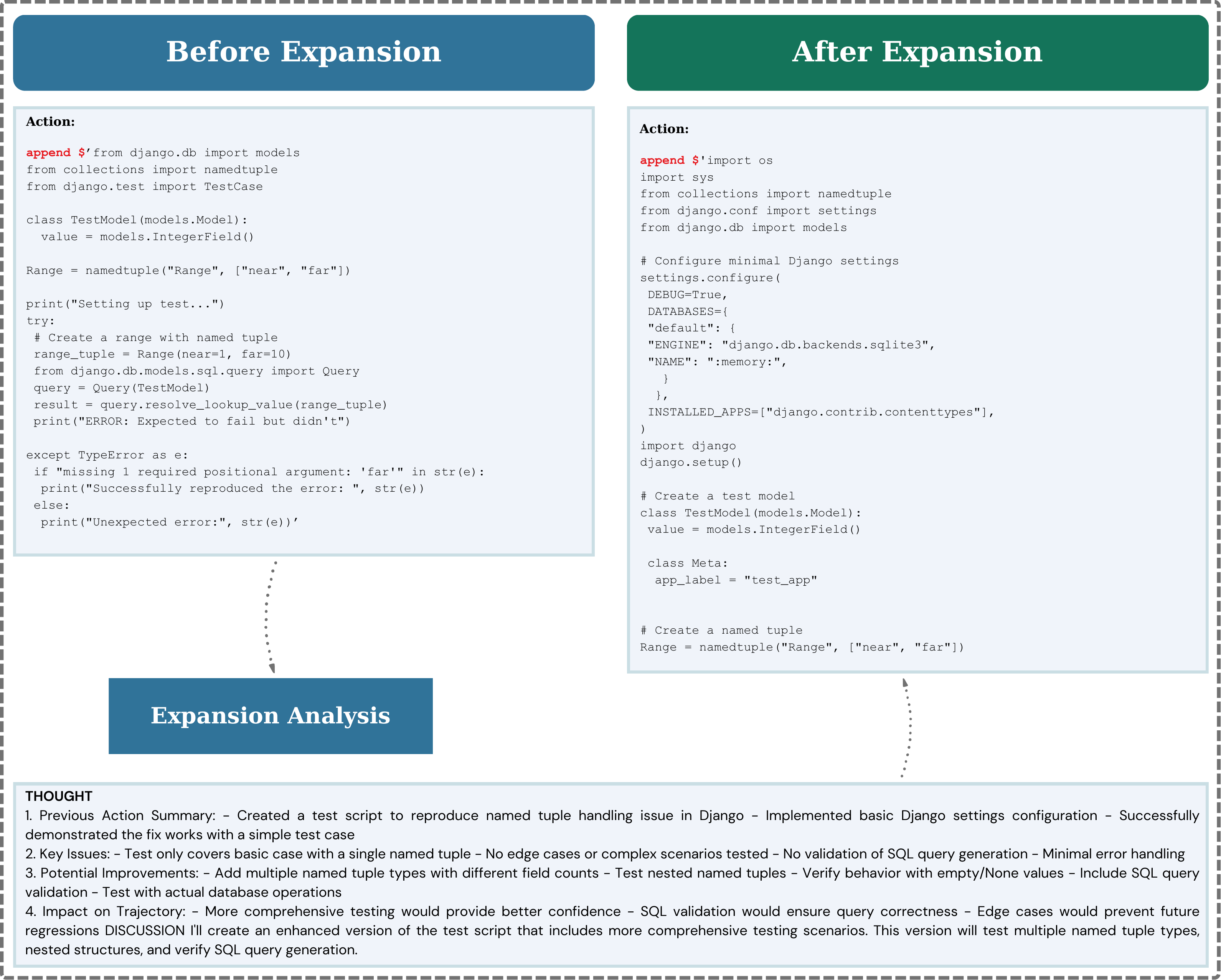}
    \caption{}
    \label{fig:append_better_rep}
\end{figure*}

\begin{figure*}[!htb]
    \centering
    \includegraphics[width=\textwidth]{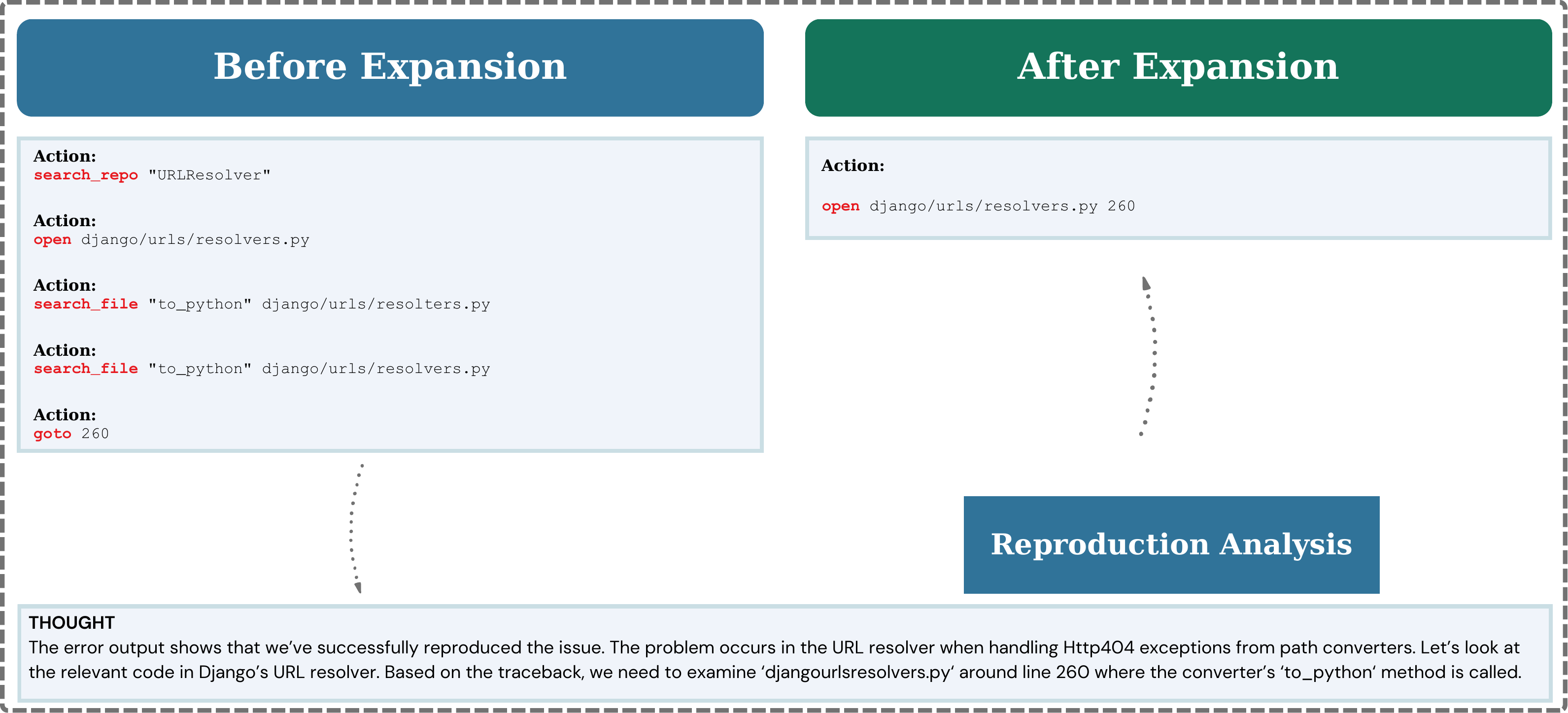}
    \caption{Append expansion leads to better reproduction script which leads to better localization.}
    \label{fig:append_better_loc}
\end{figure*}

\begin{figure*}[!htb]
    \centering
    \includegraphics[width=\textwidth]{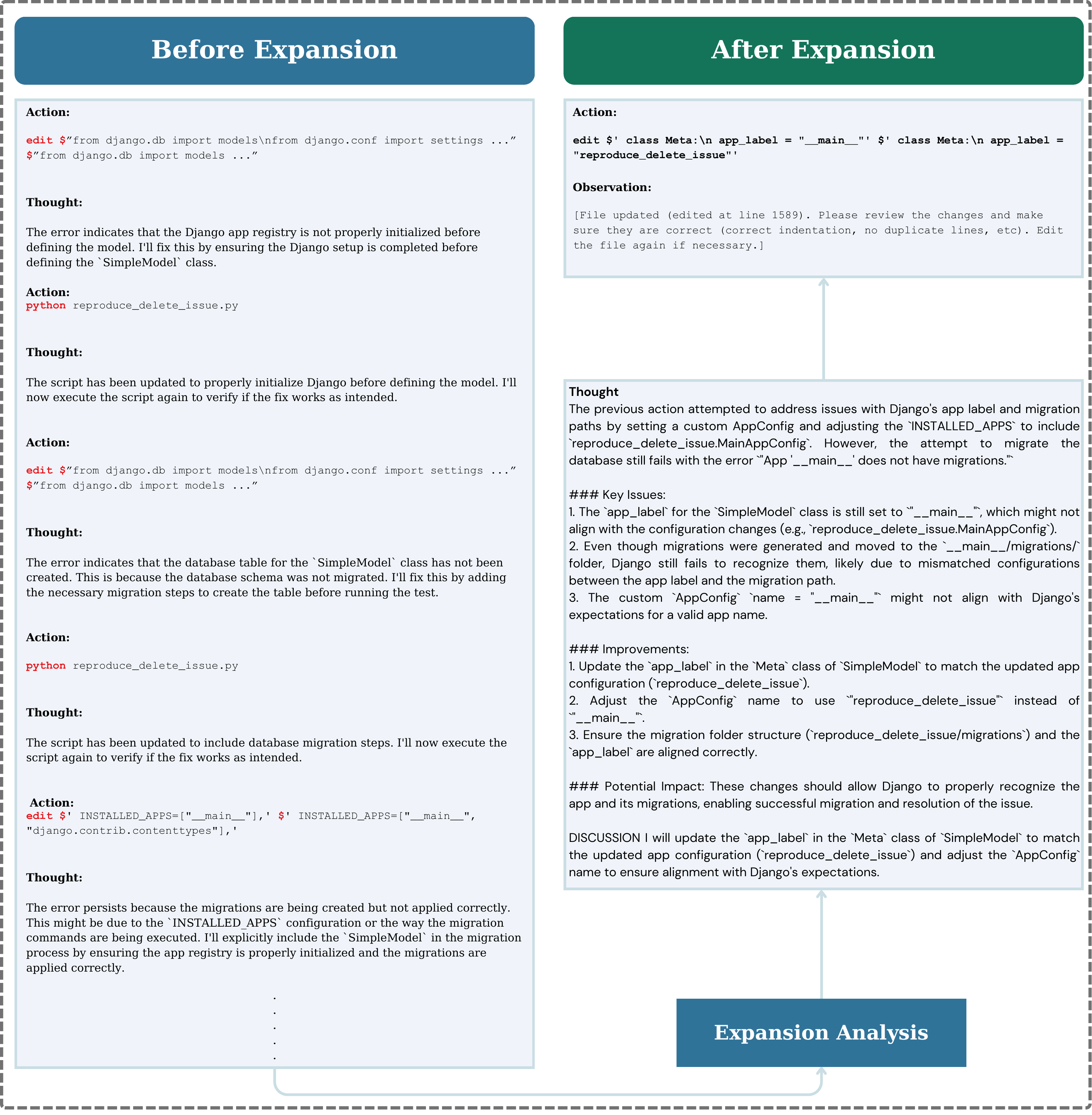}
    \caption{Expanding in edit helps the model break out of edit-python loop.}
    \label{fig:edit_python_loop}
\end{figure*}

\begin{figure*}[!htb]
    \centering
    \includegraphics[width=\textwidth]{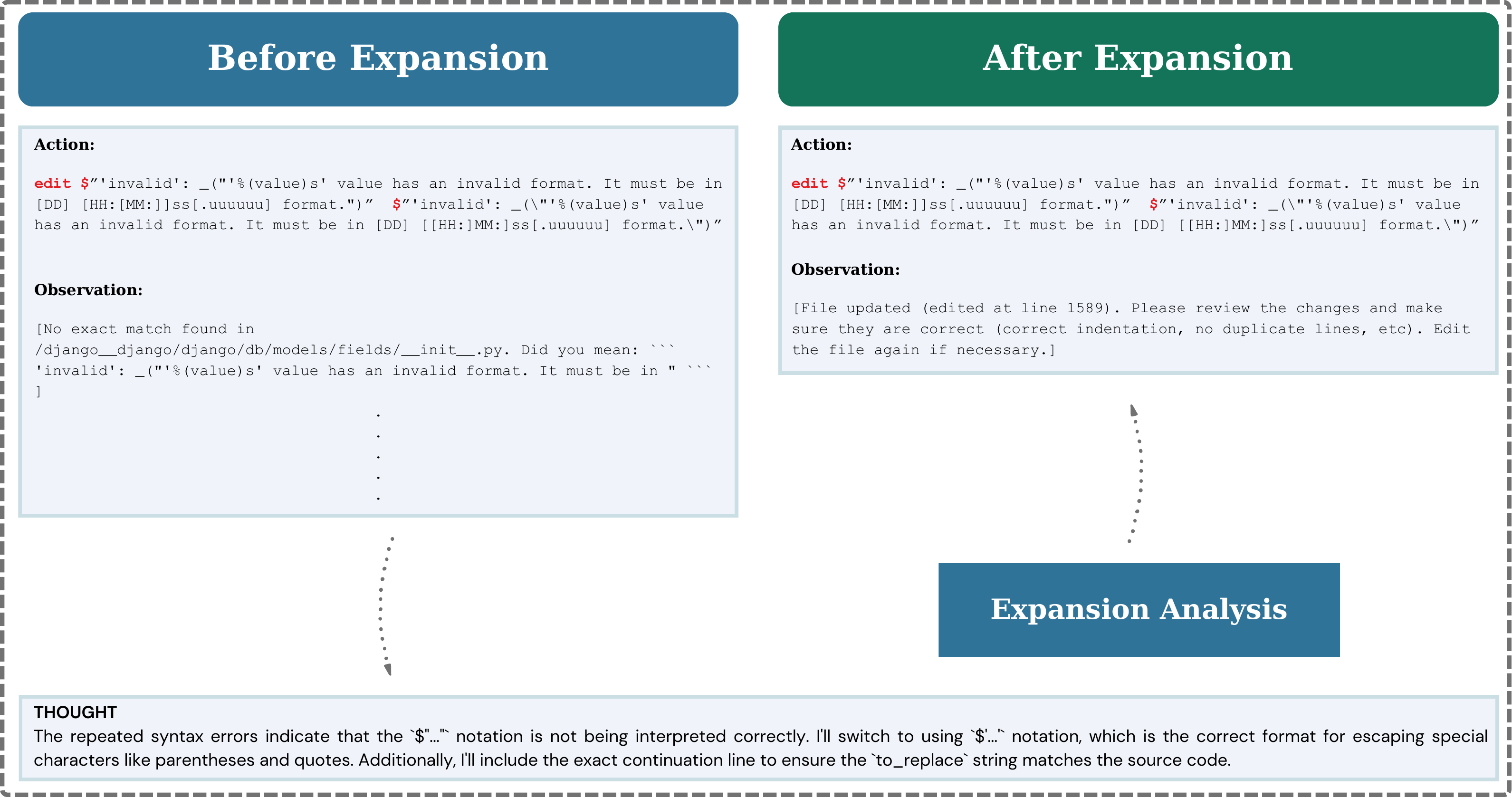}
    \caption{Model often hallucinates by repeatedly outputting the same action. The model realizes that during expansion and fixes it.}
    \label{fig:edit_loop}
\end{figure*}

\clearpage

\subsection{Trajectory Analysis Tool}

\begin{figure*}[!htb]
    \centering
    \includegraphics[width=\textwidth]{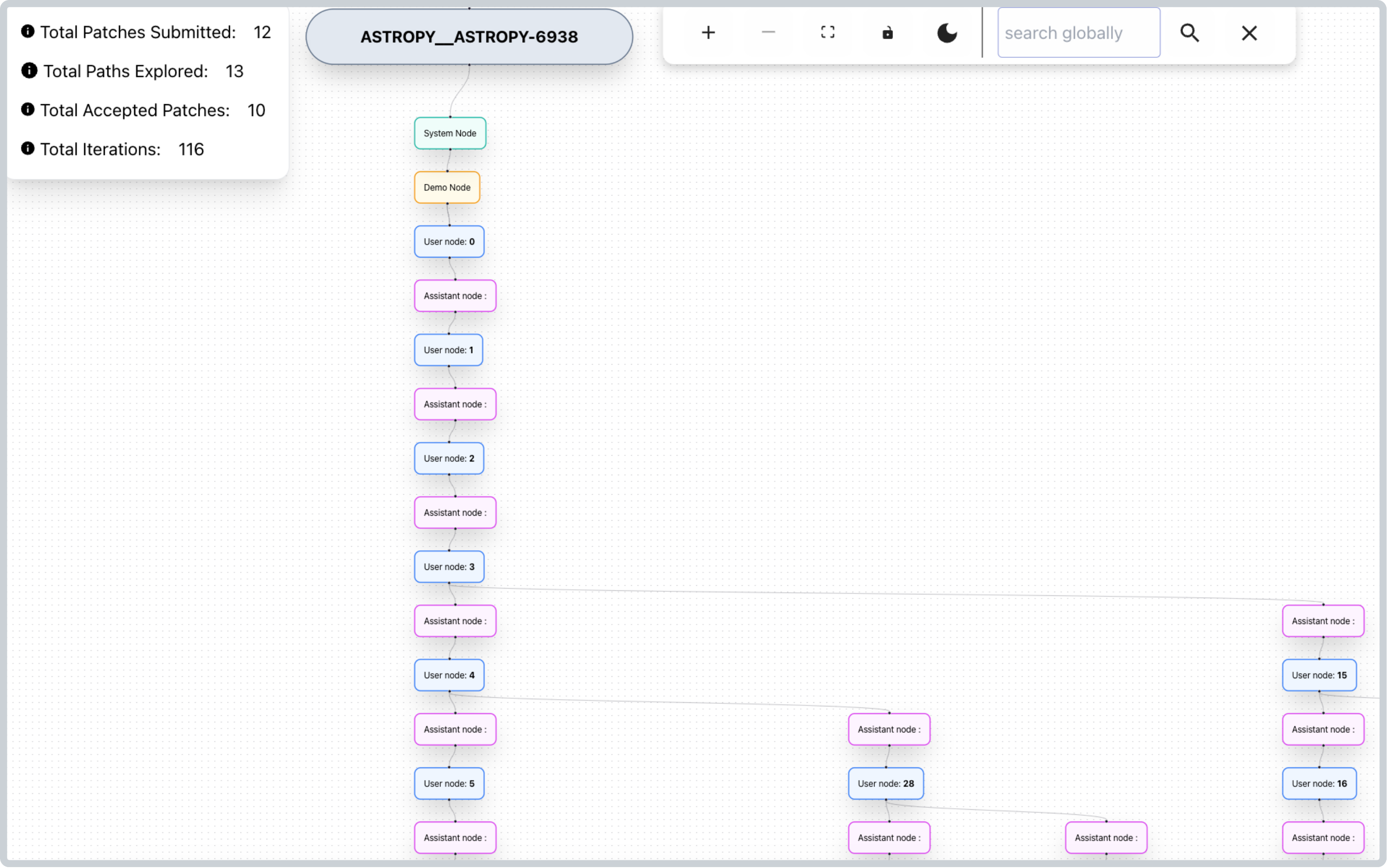}
    \caption{Overview of the entire trajectory}
    \label{fig:analysis-tool-full-graph}
\end{figure*}

\begin{figure*}[!htb]
    \centering
    \includegraphics[width=\textwidth]{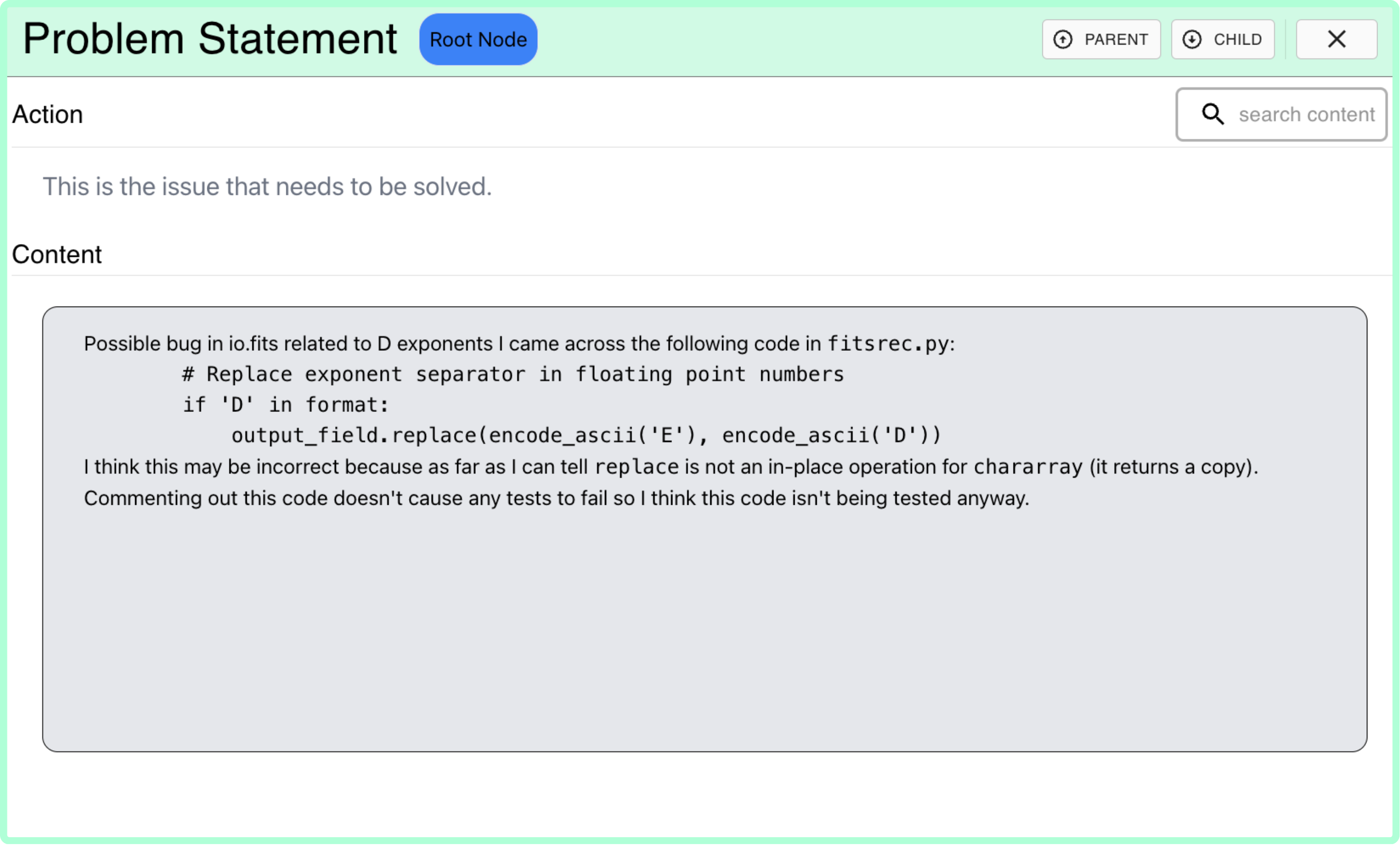}
    \caption{Problem Statement Node}
    \label{fig:analysis-tool-problem-statement}
\end{figure*}

\begin{figure*}[!htb]
    \centering
    \includegraphics[width=\textwidth]{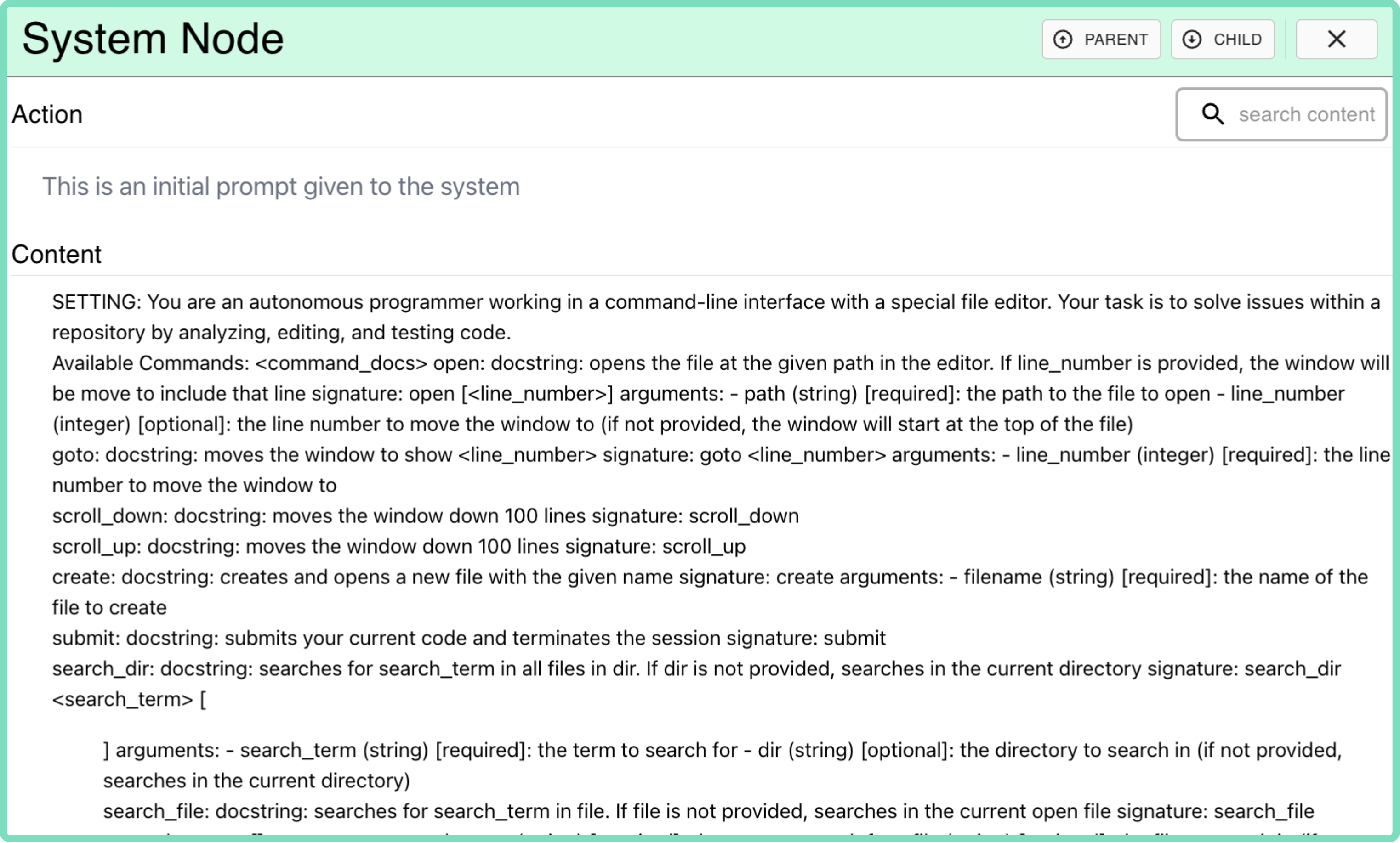}
    \caption{System Node}
    \label{fig:analysis-tool-system-node}
\end{figure*}

\begin{figure*}[!htb]
    \centering
    \includegraphics[width=\textwidth]{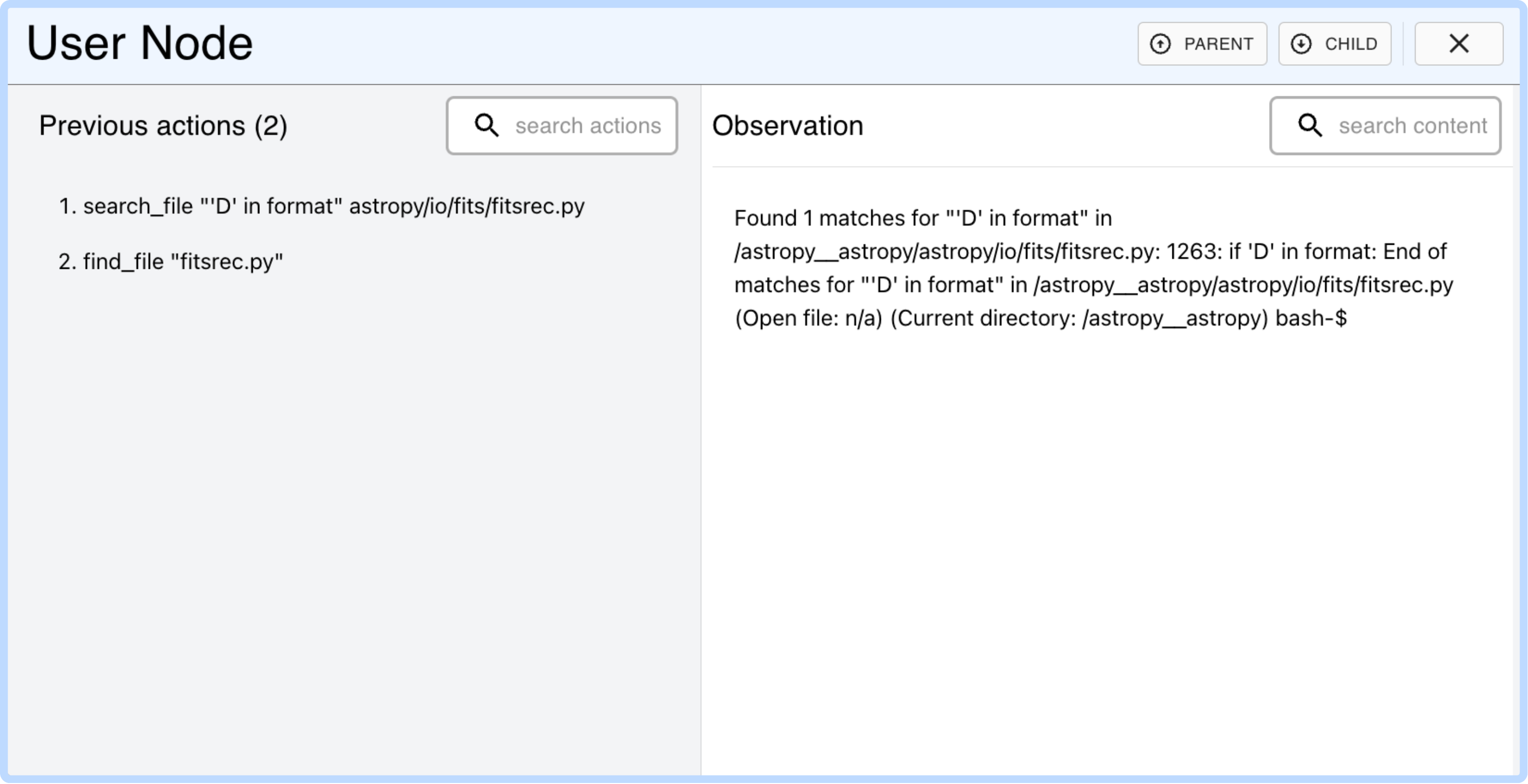}
    \caption{User Node}
    \label{fig:analysis-tool-user-node}
\end{figure*}

\begin{figure*}[!htb]
    \centering
    \includegraphics[width=\textwidth]{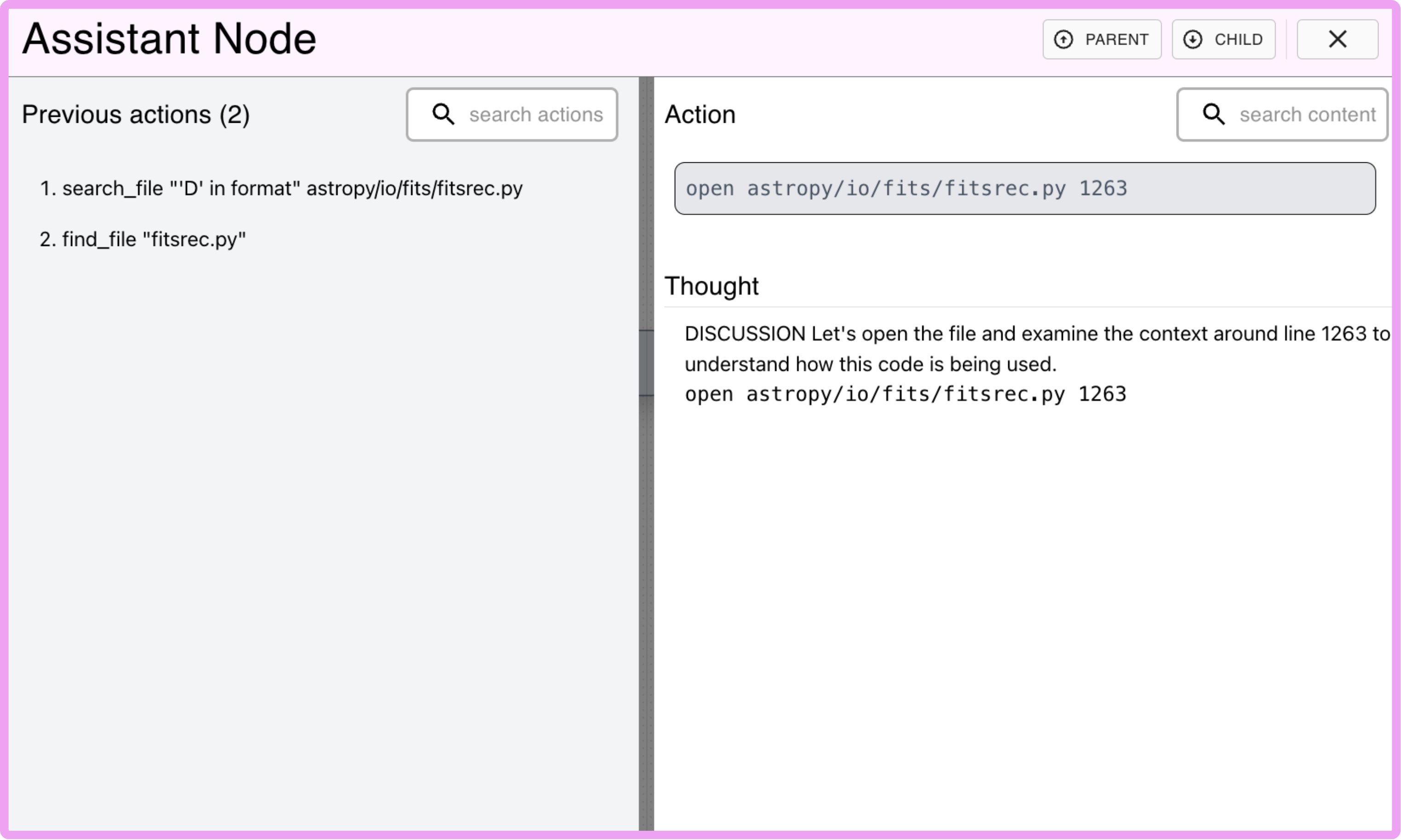}
    \caption{Assistant Node}
    \label{fig:analysis-tool-assistant-node}
\end{figure*}

\begin{figure*}[!htb]
    \centering
    \includegraphics[width=0.5\textwidth]{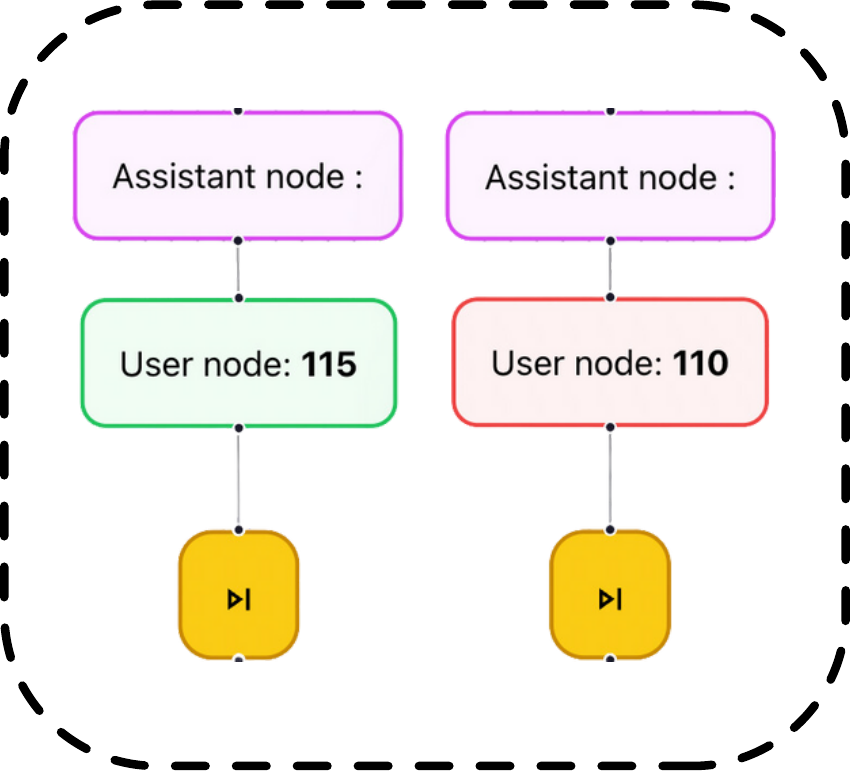}
    \caption{Assistant Node}
    \label{fig:analysis-tool-accepted-child}
\end{figure*}
\clearpage
\onecolumn
\subsection{Prompts}
\subsubsection{Backbone Agent Prompts}
This is the system prompt for the backbone agent. This contains the abstract commands, their usage and general guidelines for the agent. The agent is expected to follow these commands to interact with the environment and solve the issues in the repository.
\label{backbone_prompts}
\begin{tcolorbox}[
    breakable,
    colback=white,
    colframe={rgb:red,0;green,158;blue,115},
    title=System Prompt,
    fonttitle=\bfseries
]

SETTING: You are an autonomous programmer working in a command-line interface with a special file editor. Your task is to solve issues within a repository by analyzing, editing, and testing code.

Available Commands:

\verb|<command_docs>|
\begin{verbatim}
    open:
    docstring: opens the file at the given path in the editor. If 
    line_number is provided, the window will be move to include that line
    signature: open <path> [<line_number>]
    arguments:
      - path (string) [required]: the path to the file to open
      - line_number (integer) [optional]: the line number to move the window
      to (if not provided, the window will start at the top of the file)
  
  goto:
    docstring: moves the window to show <line_number>
    signature: goto <line_number>
    arguments:
      - line_number (integer) [required]: the line number
      to move the window to
  
  scroll_down:
    docstring: moves the window down 100 lines
    signature: scroll_down
  
  scroll_up:
    docstring: moves the window down 100 lines
    signature: scroll_up
  
  create:
    docstring: creates and opens a new file with the given name
    signature: create <filename>
    arguments:
      - filename (string) [required]: the name of the file to create
  
  submit:
    docstring: submits your current code and terminates the session
    signature: submit
  
  search_dir:
    docstring: searches for search_term in all files in dir. If dir is
    not provided, searches in the current directory
    signature: search_dir <search_term> [<dir>]
    arguments:
      - search_term (string) [required]: the term to search for
      - dir (string) [optional]: the directory to search in (if
      not provided,
      searches in the current directory)
  
  search_file:
    docstring: searches for search_term in file. If file is not provided,
    searches in the current open file
    signature: search_file <search_term> [<file>]
    arguments:
      - search_term (string) [required]: the term to search for
      - file (string) [optional]: the file to search in (if not provided,
       searches in the current open file)
  
  find_file:
    docstring: finds all files with the given name in dir. If dir is not 
    provided, searches in the current directory
    signature: find_file <file_name> [<dir>]
    arguments:
      - file_name (string) [required]: the name of the file to search for
      - dir (string) [optional]: the directory to search in (if
      not provided,
      searches in the current directory)
  
  edit:
    docstring: Replaces occurrence of $<to_replace> with $<new_content> in
     the currently open file.
    signature: edit $<to_replace> $<new_content>
    arguments:
      - to_replace (string) [required]: The text to be replaced in the file.
      - new_content (string) [required]: The new text to replace with.
  
  undo_edit:
    docstring: Reverts the last edit made to the specified file. If no
    file is provided, reverts the last edit on the currently open file.
    signature: undo_edit [file_path]
    arguments:
      - file_path (string) [optional]: The path to the file to undo the 
      last edit for.
  
  insert:
    docstring: Inserts $<content> at the given <line_number> in the 
    currently open file.
    signature: insert <line_number> $<content>
    arguments:
      - line_number (int) [required]: The line number where the content
      should be inserted.
      - content (string) [required]: The content to insert at the specified
      line number.
  
  append:
    docstring: Appends $<content> to the end of the currently open file.
    signature: append $<content>
    arguments:
      - content (string) [required]: The content to append to the end of the
      file.
  
  execute_ipython:
    docstring: Executes Python code in a persistent cell, returning its
    output. Variables persist between executions.
    signature: execute_ipython $<code>
    arguments:
      - code (string) [required]: Python code to execute in the cell.
  
  execute_server:
    docstring: To run long-lived processes such as server or daemon. It runs
    the command in the background and provides a log of the output.
    signature: execute_server <command>
    arguments:
      - command (string) [required]: Bash command to execute in the shell.
  
  
  search_repo:
    docstring: searches in the current repository with a specific function
    or class, and returns the def and ref relations for the search term.
    signature: search_repo <search_term>
    arguments:
      - search_term (string) [required]: function or class to look for in 
      the repository.
\end{verbatim}
\verb|</command_docs>|

General Guidelines:
\begin{enumerate}
    \item One command at a time: Always execute a single command and wait for feedback before proceeding.
    \item Proper indentation: When editing files, ensure correct indentation for each line.
    \item File awareness: Pay attention to the currently open file and working directory.
    \item Search functionality: Use \texttt{search\_repo} command to gather information when needed.
    \item For interactive sessions: Start it using \texttt{execute\_server} command.
\end{enumerate}

You need to format your output using two fields; discussion and command.
Your output should always include \emph{one} discussion and \emph{one} command field EXACTLY as in the following example:

DISCUSSION\\
First I'll start by using ls to see what files are in the current directory. Then maybe we can look at some relevant files to see what they look like.
\begin{verbatim}
ls -a
\end{verbatim}

\end{tcolorbox}

The following is the first user prompt for the agent. This prompt is used to describe the issue to the agent and provides special instructions regarding the use of various commands described in the system prompt.

\begin{tcolorbox}[
    breakable,
    colback=white,
    colframe={rgb:red,247;green,217;blue,188},
    title=Tool Instructions,
    fonttitle=\bfseries,
    width=\textwidth,
    before skip=12pt,
    after skip=12pt
]

\label{tool instructions}

\textbf{instance\_template:}

Here's the issue you need to address, as described in the PR:

\begin{minipage}{\dimexpr\linewidth-2\parindent}
\begin{verbatim}
<pr_description>
{issue}
</pr_description>
\end{verbatim}
\end{minipage}

You're in the repository's root directory. Can you help me implement the necessary changes to the repository so that the requirements specified in the \verb|<pr_description>| are met?

Start by creating a minimal script to replicate and verify the bug described in the issue. Ensure the bug is reproducible before making any changes. After implementing a fix, use the same script to confirm the issue is resolved. Include debugging messages, like \verb|print("Script completed successfully.")|, to indicate successful execution. The script should be focused on verification and ensuring no new errors are introduced.

Your task is to make the minimal changes to non-tests files to ensure the \verb|<pr_description>| is satisfied.

If a command fails, do not repeat it. It will not work the second time unless you modify it. Always adapt or use a different command.

\textbf{Note:} Please give only single tool call in a single step.

Follow these steps to resolve the issue:

\begin{enumerate}[leftmargin=*]
    \item Explore the repository structure to familiarize yourself with its layout.
    \item Create a script to reproduce the error and execute it using the BashTool.
    \item Edit the source code to resolve the issue, making minimal changes.
    \item Rerun your reproduce script to confirm the error is fixed.
    \item Consider edge cases and ensure your fix handles them.
\end{enumerate}

\textbf{Important Instructions for Command Usage:}

\begin{enumerate}[leftmargin=*]
    \item File Navigation:
    \begin{itemize}[leftmargin=*]
        \item Always be aware of the currently open file and the current working directory.
        \item The currently open file might be in a different directory than the working directory.
        \item Some commands, like 'create', may change the current open file.
        \item For efficient navigation to specific lines (e.g., line 583), use 'goto' instead of multiple scroll\_down commands.
    \end{itemize}

    \item Code Editing Commands (edit, append, insert):
    \begin{itemize}[leftmargin=*]
        \item If the assistant would like to add the line '\verb|        print(x)|', it must fully write the line out, with all leading spaces before the code!
        \item Prefix \verb|content| with \verb|$| to ensure the string is treated as a literal, avoiding the need for escape characters.
        \item Use \verb|$'...'| Notation: Always use \verb|$'...'| for strings in edit, append, and insert commands to correctly interpret escape sequences like \verb|\n|.
        \item Avoid \verb|$"..."| as it treats escape sequences literally.
        \item To add characters like \verb|\n| or \verb|\t| as literal strings within code, use double backslashes.
        \item Escape single or double quotes within code as \verb|\'| or \verb|\"|.
        \item Line numbers are for reference only—do not include them in \verb|content| for \verb|edit|, \verb|append|, or \verb|insert| commands.
    \end{itemize}

    \item Edit Command:
    \begin{itemize}[leftmargin=*]
        \item The \verb|to_replace| argument must exactly match the existing source code.
        \item Select the minimal number of lines necessary to uniquely identify the \verb|content|.
        \item Prefix \verb|to_replace| and \verb|new_content| with \verb|$|.
        \item Ensure \verb|new_content| includes correct indentation.
        \item To remove lines, set \verb|new_content| to an empty string.
        \item Note that \verb|to_replace| and \verb|new_content| must be different.
    \end{itemize}

    \item Insert Command:
    \begin{itemize}[leftmargin=*]
        \item Specify the exact line number for insertion.
        \item This command will not modify content before or after the specified line.
    \end{itemize}

    \item Append Command:
    \begin{itemize}[leftmargin=*]
        \item Use \verb|append| to add content to the end of a file.
        \item This is ideal after a \verb|create_file| command.
    \end{itemize}

    \item Search Command:
    \begin{itemize}[leftmargin=*]
        \item \verb|search_repo| searches the current repository for specified functions or classes.
        \item It provides definition (def) and reference (ref) relationships for the search term.
        \item \verb|search_term| is the function or class name to search for.
    \end{itemize}

    \item Execute\_server Command:
    \begin{itemize}[leftmargin=*]
        \item Use the \verb|execute_server| command to run a server or process in the background.
        \item Usage: \verb|execute_server 'Your_Command'|. Make sure to use quotes.
        \item \verb|execute_server get_logs|: Retrieves the last 100 lines of the server / process logs.
        \item \verb|execute_server stop|: Stops the background Bash server process.
    \end{itemize}
\end{enumerate}

\vspace{1em}
(Open file: \verb|{open_file}|)\\
(Current directory: \verb|{working_dir}|)\\
\verb|bash-$|

\end{tcolorbox}

Here is a demonstration prompt for the backbone agent. This prompt is used to demonstrate the correct usage of the commands and the interface to the agent.

\begin{tcolorbox}[
    breakable,
    colframe={rgb:red,124;green,161;blue,204},
    title=Demonstration Template,
    fonttitle=\bfseries,
    width=\textwidth,
    before skip=12pt,
    after skip=12pt
]

    Here is a demonstration of how to correctly accomplish this task.
    It is included to show you how to correctly use the interface.
    You do not need to follow exactly what is done in the demonstration.
    \begin{verbatim}
--- DEMONSTRATION ---
We're currently solving the following issue within our repository.
Here's the issue text:
ISSUE:
Fix: TimeDelta Precision Errors
Use microsecond integer arithmetic to fix high precision
timedelta errors.

Fixes #1865
Let's first start by reproducing the results of the issue. The issue
includes some example code for reproduction, which we can use. We'll
create a new file called `reproduce.py` and paste the example code
into it.
```
create reproduce.py
```
[File: /marshmallow-code__marshmallow/reproduce.py (1 lines total)]
1:
(Open file: /marshmallow-code__marshmallow/reproduce.py)
(Current directory: /marshmallow-code__marshmallow)
bash-$
Now let's paste in the example code from the issue.\n
```
append $'from marshmallow.fields import
TimeDelta\nfrom datetime import timedelta\n\ntd_field =
TimeDelta(precision="milliseconds")\n\nobj = dict()\nobj["td_field"] =
timedelta(milliseconds=345)\n\n
print(td_field.serialize("td_field", obj))'
```
[File: /marshmallow-code__marshmallow/reproduce.py
(9 lines total after edit)]
(this is the beginning of the file)
1:from marshmallow.fields import TimeDelta
2:from datetime import timedelta
3:
4:td_field = TimeDelta(precision="milliseconds")
5:
6:obj = dict()
7:obj["td_field"] = timedelta(milliseconds=345)
8:
9:print(td_field.serialize("td_field", obj))
(this is the end of the file)
[File updated (edited at line 1). Please review the changes and
make sure they are correct (correct indentation, no duplicate lines,
etc). Edit the file again if necessary.]
(Open file: /marshmallow-code__marshmallow/reproduce.py)
(Current directory: /marshmallow-code__marshmallow)
bash-$
Now let's run the code to see if we see the same output as the issue.\n
```
python reproduce.py
```
344
(Open file: /marshmallow-code__marshmallow/reproduce.py)
(Current directory: /marshmallow-code__marshmallow)
bash-$
We are indeed seeing the same output as the issue.The issue suggests
that we should look at line 1474 of the `fields.py` file to see
if there is a rounding issue. Let's open the `fields.py` file and
navigate to line 1474 to see the relevant code. First we should
find the `fields.py` file in the repository. It is likely to be in
the `src/marshmallow` directory, but we
should check to be sure.\n
```
ls -F
```
AUTHORS.rst
CHANGELOG.rst
CODE_OF_CONDUCT.md
CONTRIBUTING.rst
LICENSE
MANIFEST.in
NOTICE
README.rst
RELEASING.md
azure-pipelines.yml
docs/
examples/
performance/
pyproject.toml
reproduce.py
setup.cfg
setup.py
src/
tests/
tox.ini
(Open file: /marshmallow-code__marshmallow/reproduce.py)
(Current directory: /marshmallow-code__marshmallow)
bash-$
It looks like the `src` directory is present, which
suggests that the `fields.py` file is likely to be
in the `src` directory. Let's use find_file to see where it is.\n
```
find_file "fields.py" src
```
Found 1 matches for "fields.py" in /marshmallow-code__marshmallow/src:
/marshmallow-code__marshmallow/src/marshmallow/fields.py
(Open file: /marshmallow-code__marshmallow/reproduce.py)
(Current directory: /marshmallow-code__marshmallow)
bash-$
It looks like the `fields.py` file is present in the
`./src/marshmallow/` directory. The issue also points to a
specific URL with line number 1474.
We should navigate to that line in fields.py to see the relevant
code for the `TimeDelta` serialization.\n
```
open src/marshmallow/fields.py 1474
```
[File: /marshmallow-code__marshmallow/src/marshmallow/fields.py
(1997 lines total)]
(1458 more lines above)
1459:            self.WEEKS,
1460:        )
1461:
1462:        if precision not in units:
1463:            msg = 'The precision must be {} or "{}".'.format(
1464:                ", ".join([f'"{each}"' for each in units[:-1]]),
                        units[-1]
1465:            )
1466:            raise ValueError(msg)
1467:
1468:        self.precision = precision
1469:        super().__init__(**kwargs)
1470:
1471:    def _serialize(self, value, attr, obj, **kwargs):
1472:        if value is None:
1473:            return None
1474:        base_unit = dt.timedelta(**{self.precision: 1})
1475:        return int(value.total_seconds() / base_unit.total_seconds())
1476:
1477:    def _deserialize(self, value, attr, data, **kwargs):
1478:        try:
1479:            value = int(value)
1480:        except (TypeError, ValueError) as error:
1481:            raise self.make_error("invalid") from error
1482:
1483:        kwargs = {self.precision: value}
1484:
1485:        try:
1486:            return dt.timedelta(**kwargs)
1487:        except OverflowError as error:
1488:            raise self.make_error("invalid") from error
1489:
1490:
1491:class Mapping(Field):
1492:    """An abstract class for objects with key-value pairs.
1493:
1494:    :param keys: A field class or instance for dict keys.
1495:    :param values: A field class or instance for dict values.
1496:    :param kwargs: The same keyword arguments that :class:`Field`
            receives.
1497:
1498:    .. note::
1499:        When the structure of nested data is not known,
                you may omit the
1500:        `keys` and `values` arguments to prevent content validation.
1501:
1502:    .. versionadded:: 3.0.0rc4
1503:    """
1504:
1505:    mapping_type = dict
1506:
1507:    #: Default error messages.
1508:    default_error_messages = {"invalid": "Not a valid mapping type."}
1509:
1510:    def __init__(
1511:        self,
1512:        keys: typing.Optional[typing.Union[Field, type]] = None,
1513:        values: typing.Optional[typing.Union[Field, type]] = None,
1514:        **kwargs,
1515:    ):
1516:        super().__init__(**kwargs)
1517:        if keys is None:
1518:            self.key_field = None
1519:        else:
1520:            try:
1521:                self.key_field = resolve_field_instance(keys)
1522:            except FieldInstanceResolutionError as error:
1523:                raise ValueError(
1524:                    '"keys" must be a subclass or instance of '
1525:                    "marshmallow.base.FieldABC."
1526:                ) from error
1527:
1528:        if values is None:
1529:            self.value_field = None
1530:        else:
1531:            try:
1532:                self.value_field = resolve_field_instance(values)
1533:            except FieldInstanceResolutionError as error:
1534:                raise ValueError(
1535:                    '"values" must be a subclass or instance of '
1536:                    "marshmallow.base.FieldABC."
1537:                ) from error
1538:            if isinstance(self.value_field, Nested):
1539:                self.only = self.value_field.only
1540:                self.exclude = self.value_field.exclude
1541:
1542:    def _bind_to_schema(self, field_name, schema):
1543:        super()._bind_to_schema(field_name, schema)
1544:        if self.value_field:
1545:            self.value_field = copy.deepcopy(self.value_field)
1546:            self.value_field._bind_to_schema(field_name, self)
1547:        if isinstance(self.value_field, Nested):
1548:            self.value_field.only = self.only
1549:            self.value_field.exclude = self.exclude
1550:        if self.key_field:
1551:            self.key_field = copy.deepcopy(self.key_field)
1552:            self.key_field._bind_to_schema(field_name, self)
1553:
1554:    def _serialize(self, value, attr, obj, **kwargs):
1555:        if value is None:
1556:            return None
1557:        if not self.value_field and not self.key_field:
1558:            return self.mapping_type(value)
(439 more lines below)
(Open file: /marshmallow-code__marshmallow/src/marshmallow/fields.py)
(Current directory: /marshmallow-code__marshmallow)
bash-$
We are now looking at the relevant section of the `fields.py` file
where the `TimeDelta` serialization occurs. The issue suggests that
there is a rounding problem near line 1474. The current implementation
uses integer division, which truncates the decimal part and can lead
to incorrect rounding on line 1475.To fix this issue, we should ensure
that the division results in a float, and then round the result to
the nearest integer before converting it to an int. This should
preserve the milliseconds precision as expected. Let's make the
necessary edit to the code.\n
```
edit $'        return int(value.total_seconds() /
base_unit.total_seconds())'
$'        return int(round(value.total_seconds() /
base_unit.total_seconds()))'
```
[File: /marshmallow-code__marshmallow/src/marshmallow/fields.py
(1997 lines total after edit)]
(1424 more lines above)
1425:
1426:    :param precision: Influences how the integer is interpreted
                            during
1427:        (de)serialization. Must be 'days', 'seconds', 'microseconds',
1428:        'milliseconds', 'minutes', 'hours' or 'weeks'.
1429:    :param kwargs: The same keyword arguments that :class:`Field`
                        receives.
1430:
1431:    .. versionchanged:: 2.0.0
1432:        Always serializes to an integer value to avoid rounding
                errors.
1433:        Add `precision` parameter.
1434:    """
1435:
1436:    DAYS = "days"
1437:    SECONDS = "seconds"
1438:    MICROSECONDS = "microseconds"
1439:    MILLISECONDS = "milliseconds"
1440:    MINUTES = "minutes"
1441:    HOURS = "hours"
1442:    WEEKS = "weeks"
1443:
1444:    #: Default error messages.
1445:    default_error_messages = {
1446:        "invalid": "Not a valid period of time.",
1447:        "format": "{input!r} cannot be formatted as a timedelta.",
1448:    }
1449:
1450:    def __init__(self, precision: str = SECONDS, **kwargs):
1451:        precision = precision.lower()
1452:        units = (
1453:            self.DAYS,
1454:            self.SECONDS,
1455:            self.MICROSECONDS,
1456:            self.MILLISECONDS,
1457:            self.MINUTES,
1458:            self.HOURS,
1459:            self.WEEKS,
1460:        )
1461:
1462:        if precision not in units:
1463:            msg = 'The precision must be {} or "{}".'.format(
1464:                ", ".join([f'"{each}"' for each in units[:-1]]),
                        units[-1]
1465:            )
1466:            raise ValueError(msg)
1467:
1468:        self.precision = precision
1469:        super().__init__(**kwargs)
1470:
1471:    def _serialize(self, value, attr, obj, **kwargs):
1472:        if value is None:
1473:            return None
1474:        base_unit = dt.timedelta(**{self.precision: 1})
1475:        return int(round(value.total_seconds() / 
                            base_unit.total_seconds()))
1476:
1477:    def _deserialize(self, value, attr, data, **kwargs):
1478:        try:
1479:            value = int(value)
1480:        except (TypeError, ValueError) as error:
1481:            raise self.make_error("invalid") from error
1482:
1483:        kwargs = {self.precision: value}
1484:
1485:        try:
1486:            return dt.timedelta(**kwargs)
1487:        except OverflowError as error:
1488:            raise self.make_error("invalid") from error
1489:
1490:
1491:class Mapping(Field):
1492:    """An abstract class for objects with key-value pairs.
1493:
1494:    :param keys: A field class or instance for dict keys.
1495:    :param values: A field class or instance for dict values.
1496:    :param kwargs: The same keyword arguments that :class:`Field`
        receives.
1497:
1498:    .. note::
1499:        When the structure of nested data is not known, you may
1500:        omit the `keys` and `values` arguments to prevent content
                validation.
1501:
1502:    .. versionadded:: 3.0.0rc4
1503:    """
1504:
1505:    mapping_type = dict
1506:
1507:    #: Default error messages.
1508:    default_error_messages = {"invalid": "Not a valid mapping type."}
1509:
1510:    def __init__(
1511:        self,
1512:        keys: typing.Optional[typing.Union[Field, type]] = None,
1513:        values: typing.Optional[typing.Union[Field, type]] = None,
1514:        **kwargs,
1515:    ):
1516:        super().__init__(**kwargs)
1517:        if keys is None:
1518:            self.key_field = None
1519:        else:
1520:            try:
1521:                self.key_field = resolve_field_instance(keys)
1522:            except FieldInstanceResolutionError as error:
1523:                raise ValueError(
1524:                    '"keys" must be a subclass or instance of '
1525:                    "marshmallow.base.FieldABC."
(472 more lines below)
[File updated (edited at line 1475). Please review the changes and make
sure they are correct (correct indentation, no duplicate lines, etc).
Edit the file again if necessary.]
(Open file: /marshmallow-code__marshmallow/src/marshmallow/fields.py)
(Current directory: /marshmallow-code__marshmallow)
bash-$
The code has been updated to use the `round` function, which should fix
the rounding issue. Before submitting the changes, it would be prudent
to run the reproduce.py code again to ensure that this change has
the desired effect. I see that "Current directory" is still
`/marshmallow-code__marshmallow`, so we can just run `reproduce.py`
directly.\n
```
python reproduce.py
```
345
(Open file: /marshmallow-code__marshmallow/src/marshmallow/fields.py)
(Current directory: /marshmallow-code__marshmallow)
bash-$
The output has changed from 344 to 345, which suggests that the rounding
issue has been fixed. Let's remove the reproduce.py file since it is no
longer needed.\n
```
rm reproduce.py
```
Your command ran successfully and did not produce any output.
(Open file: /marshmallow-code__marshmallow/src/marshmallow/fields.py)
(Current directory: /marshmallow-code__marshmallow)
bash-$
rm doesn't have any output when it deletes successfully, so that must
have worked. Now that we have fixed the issue, we should submit our
changes to the code base. Let's submit the changes using the `submit`
command.\n
```
submit
```
--- END OF DEMONSTRATION ---
\end{verbatim}

\end{tcolorbox}

Here is a template for the next step prompt. Observations from the environment are provided to the agent using this template.

\begin{tcolorbox}[
    breakable,
    colframe=observationframe,
    colback=white,
    title=Next Step Template,
    fonttitle=\bfseries,
    width=\textwidth,
    before skip=12pt,
    after skip=12pt
]

\begin{verbatim}
{observation}
(Open file: {open_file})
(Current directory: {working_dir})
bash-$
\end{verbatim}
\end{tcolorbox}

This template is used to indicate that the last command the agent executed was successful and did not produce any output.

\begin{tcolorbox}[
    breakable,
    colframe={rgb:red,0.043;green,0.322;blue,0.580},
    colback=white,
    title=Next Step no Output Template,
    fonttitle=\bfseries,
    width=\textwidth,
    before skip=12pt,
    after skip=12pt
]

\begin{verbatim}
Your command ran successfully and did not produce any output.
(Open file: {open_file})
(Current directory: {working_dir})
bash-$
\end{verbatim}

\end{tcolorbox}

This template is specifically tailored for the search\_repo command. It is used to display the search results to the agent.

\begin{tcolorbox}[
    breakable,
    colframe={rgb:red,247;green,217;blue,188},
    colback=white,
    title=Search Results,
    fonttitle=\bfseries,
    width=\textwidth,
    before skip=12pt,
    after skip=12pt
]

\begin{minipage}{\dimexpr\linewidth-2\parindent}
\begin{verbatim}
Your command ran successfully and produced the following related
functions/classes for {search_term}:
For each item, `fname` denotes the source file, `line`
denotes the line number, `kind` means whether it
is definition or reference, and `info` contains the specific content.
{codegraph_context}
(Open file: {open_file})
(Current directory: {working_dir})
bash-$
\end{verbatim}
\end{minipage}
\end{tcolorbox}

\subsubsection{Expansion Prompts}

This section provides templates for the expansion prompts. These prompts are used to guide the agent in suggesting improved alternate actions using execution feedback from the previous trajectory.

\begin{tcolorbox}[
    breakable,
    colframe={rgb:red,155;green,19;blue,27},
    colback=white,
    title=Edit Expansion Template,
    fonttitle=\bfseries,
    width=\textwidth,
    before skip=12pt,
    after skip=12pt
]

\begin{verbatim}
You will be given information about a previous action and its trajectory.
Your goal is to suggest a refined or alternative action that better resolves
the issue at hand.
Here is the information about the previous modification:

Previous action:
<previous_action>
{action}
</previous_action>

Trajectory after the action:
<previous_trajectory>
{prev_traj}
</previous_trajectory>

Instructions:
1. Analyze the previous action and its trajectory.
2. Suggest a replacement action that improves upon the previous one.
3. Focus on refining the current edit, modifying different sections,
or making small insertions as needed.
4. Keep your suggestion concise and directly related to the file
modification.

Before providing your final suggestion, wrap your analysis
process in <analysis> tags. In this analysis:
1. Summarize the previous action and its trajectory
2. Identify the key issues or shortcomings in the previous action
3. List potential improvements or alternative approaches
4. Consider how these changes might affect the trajectory

You need to format your output using three fields; analysis,
discussion and command.
\end{verbatim}

\end{tcolorbox}

\begin{tcolorbox}[
    breakable,
    colframe={rgb:red,155;green,19;blue,27},
    colback=white,
    title=Insert Expansion Template,
    fonttitle=\bfseries,
    width=\textwidth,
    before skip=12pt,
    after skip=12pt
]

\begin{verbatim}
You will be given information about a previous action and its trajectory.
Your goal is to suggest a single, concise improvement that replaces the
previous action. Here's the information about the previous modification:

Previous action:
<action>
{action}
</action>

Trajectory after the action:
<prev_traj>
{prev_traj}
</prev_traj>

Your task is to analyze this information and suggest one improvement.
This improvement should replace the previous action, not be a next step.
Focus on one of these approaches:
1. A different insertion with varied content
2. An insertion in a new location
3. Editing existing content for a more effective resolution

Before providing your final suggestion, wrap your analysis process in
<analysis> tags. In this analysis:
1. Summarize the previous action and its trajectory
2. Identify the key issues or shortcomings in the previous action
3. List potential improvements or alternative approaches
4. Consider how these changes might affect the trajectory

You need to format your output using three fields; analysis,
discussion and command.
\end{verbatim}

\end{tcolorbox}

\begin{tcolorbox}[
    breakable,
    colframe={rgb:red,155;green,19;blue,27},
    colback=white,
    title=Append Expansion Template,
    fonttitle=\bfseries,
    width=\textwidth,
    before skip=12pt,
    after skip=12pt
]

\begin{verbatim}
Your goal is to suggest alternative content for appending to a file,
based on a previous action and its outcome.
Here's the information about the previous operation:

<previous_action>
{action}
</previous_action>

<previous_trajectory>
{prev_traj}
</previous_trajectory>

Your task is to suggest a replacement for the previous append action, not to
provide the next action in the sequence. The reproduction script you've
written may lack completeness on its own. Would you like to review
it and write a more comprehensive version of the script, incorporating
the context of the previous trajectory?

1. Analyze the previous action:
    - What specific content was appended?
    - What was the likely purpose of this content?

2. Brainstorm at least three alternative content ideas:
    - Describe each alternative and how it differs from the original.
    - Number each alternative for easy reference.

3. Evaluate each alternative:
    - How does it potentially improve exploration?
    - What new insights might it provide?

4. Select the best alternative:
    - Which option do you think is most promising?
    - Justify your choice in 1-2 sentences.

Before providing your final suggestion, wrap your analysis process in
<analysis> tags. In this analysis:
1. Summarize the previous action and its trajectory
2. Identify the key issues or shortcomings in the previous action
3. List potential improvements or alternative approaches
4. Consider how these changes might affect the trajectory

You need to format your output using three fields; analysis, discussion and
command.
\end{verbatim}

\end{tcolorbox}

\begin{tcolorbox}[
    breakable,
    colframe={rgb:red,155;green,19;blue,27},
    colback=white,
    title=Submit Expansion Template,
    fonttitle=\bfseries,
    width=\textwidth,
    before skip=12pt,
    after skip=12pt
]

\begin{verbatim}
You are about to submit the changes. Have you double-checked that your
changes don't affect other test cases or have any unintended consequences or
completely fix the issue? Please review once more before submitting.
\end{verbatim}

\end{tcolorbox}

\begin{tcolorbox}[
    breakable,
    colframe={rgb:red,155;green,19;blue,27},
    title=Create Expansion Template,
    fonttitle=\bfseries,
    width=\textwidth,
    before skip=12pt,
    after skip=12pt
]

\begin{verbatim}
Before trying to reproduce the bug, let's first try to localize the issue,
we can test the issue after the fix.
\end{verbatim}

\end{tcolorbox}

\begin{tcolorbox}[
    breakable,
    colframe={rgb:red,155;green,19;blue,27},
    colback=white,
    title=Critic Prompt Template,
    fonttitle=\bfseries,
    width=\textwidth,
    before skip=12pt,
    after skip=12pt
]

\begin{verbatim}
You are an AI system tasked with selecting the best alternative action to
replace a previously executed action in a process or workflow. Your goal is
to evaluate the given alternatives and choose the most effective
replacement.
Here is the previously executed action:
<previous_action>

{previous_action}
</previous_action>

Here is the list of alternative actions to consider:
<alternative_actions>
{actions}
</alternative_actions>

Instructions:
1. Evaluate each action in the list of alternative actions based on the
following criteria:
    a. It must be different from the previous action.
    b. It should replace the previous action, not be implemented after it.
    c. It should be more effective than the previous action.

2. Analyze each action inside <action_analysis> tags, following this
structure:
    - List each action with a number.
    - For each action, explicitly state whether it meets each of the three
    criteria.
    - Provide a brief explanation for why the action does or doesn't meet
    each criterion.
    - If the action meets all criteria, give it a numerical
    effectiveness score (1-10).

3. After evaluating all actions, select the best one that meets all the
criteria and is the most effective replacement for the previous action.

4. Provide the index of the best action using <best_action_index>
tags starting from 0.

Example output format:
<action_analysis>
[All actions analysis one by one]
</action_analysis>

<best_action_index>[Your selected best action index]</best_action_index>
\end{verbatim}

\end{tcolorbox}

\subsubsection{Trajectory Selection Prompts}

This prompt is used to get the best patch among all the patches generated by the agent. It two rubrics to evaluate the patches and select the best one.

\begin{tcolorbox}[
    breakable,
    colback=white,
    colframe={rgb:red,166;green,77;blue,122},
    title=Patch Analysis Guidelines,
    fonttitle=\bfseries
]

SETTING: You are an expert software engineering evaluator analyzing patches for GitHub issues. Your task is to evaluate and select the most effective solution patch.

Evaluation Criteria:

\begin{verbatim}
1. Bug Fixing Score (0-2):
   0: Incorrect changes that won't fix the issue
   1: Partially correct changes (might fix some cases)
   2: Correct changes that fully fix the issue

2. Regression Risk (0-2):
   0: High regression risk
   1: Moderate regression risk
   2: Low regression risk
\end{verbatim}

Analysis Format:

\verb|<patch_analysis>|
\begin{verbatim}
  <patch_number>[Number]</patch_number>
  <bug_fixing_analysis>
    [Analysis of fix approach]
    <score>[0-2]</score>
  </bug_fixing_analysis>
  <regression_risk_analysis>
    [Analysis of risks]
    <score>[0-2]</score>
  </regression_risk_analysis>
</patch_analysis>
\end{verbatim}

Key Considerations:
\begin{enumerate}
    \item Core issue resolution effectiveness
    \item Potential regression impacts
    \item Edge case handling
    \item Implementation quality
\end{enumerate}

Your analysis should include:
\begin{itemize}
    \item Detailed patch changes evaluation
    \item Side-by-side comparison
    \item Edge case consideration
    \item Independent assessment
\end{itemize}

Final Output Format:

\verb|<best_patch>|[Selected patch number]\verb|</best_patch>|

\end{tcolorbox}

This prompt is used to critique a generated patch based on the output of running test cases after applying the patch.

\begin{tcolorbox}[
    breakable,
    colback=white,
    colframe={rgb:red,166;green,77;blue,122},
    title=Critique Generation Template,
    fonttitle=\bfseries
]

SETTING: You are an expert software engineer evaluating a proposed patch for a GitHub issue. Your task is to analyze and critique the effectiveness of the solution.

Evaluation Steps:

\begin{enumerate}
    \item Examine patch content in:
    \verb|<patch>|[patch content]\verb|</patch>|
    
    \item Review issue details in:
    \verb|<github_issue>|[issue description]\verb|</github_issue>|
    
    \item Consider patch status in:
    \verb|<patch_status>|[status details]\verb|</patch_status>|
    
    \item Apply scoring criteria:
    \begin{verbatim}
    Bug Fixing Score (0-2):
    0: Incorrect changes
    1: Partially correct changes
    2: Correct changes

    Regression Risk (0-2):
    0: High regression risk
    1: Moderate regression risk
    2: Low regression risk
    \end{verbatim}
    
    \item Review test results in:
    \verb|<bug_fixing_tests>|[test results]\verb|</bug_fixing_tests>|
    \verb|<regression_risk_tests>|[risk results]\verb|</regression_risk_tests>|
\end{enumerate}

Analysis Format:

\verb|<evaluation>|
\begin{verbatim}
[Detailed analysis including:
- Relevant patch/issue quotes
- Solution explanation
- Effectiveness assessment
- Risk-benefit analysis]
\end{verbatim}
\verb|</evaluation>|

Critique Format:

\verb|<critique>|
\begin{verbatim}
[Concise (<100 words) summary focusing on:
- Key effectiveness points
- Critical impact factors
Note: Positive for solved status,
      negative for unsolved status]
\end{verbatim}
\verb|</critique>|

Important Notes:
\begin{itemize}
    \item Avoid mentioning specific test names
    \item Maintain clear, focused language
    \item Analyze as if status and test results were unknown
    \item Keep critique concise and impactful
\end{itemize}

\end{tcolorbox}

\end{document}